\documentclass[conference]{IEEEtran}
\usepackage{times}

\usepackage[numbers]{natbib}
\usepackage{multicol}
\usepackage[bookmarks=true]{hyperref}
\usepackage{graphicx}

\usepackage{multirow}
\usepackage{multicol}
\usepackage{graphicx}
\usepackage{xspace}
\usepackage{xcolor}
\usepackage{caption}
\usepackage{wrapfig}
\usepackage{bbding}  
\usepackage{pifont}  

\usepackage{bm}
\usepackage{booktabs}
\usepackage{float}
\usepackage{amsmath}  
\usepackage{amssymb}  

\usepackage[capitalise, nameinlink]{cleveref}

\usepackage{colortbl}
\definecolor{ourcolor}{HTML}{99e0eb}
\definecolor{ourblue}{HTML}{27a2c3}

\definecolor{tablecolor}{HTML}{ccf2f5} 

\definecolor{tablecolor2}{HTML}{ffcdb4}
\definecolor{citecolor}{HTML}{fe7b5b}
\definecolor{grey}{rgb}{0.9, 0.9, 0.9}
\usepackage{amssymb}

\usepackage{listings}
\definecolor{gred}{rgb}{0.859,0.267,0.216}
\definecolor{ggreen}{rgb}{0.059,0.616,0.345}

\definecolor{deepblue}{HTML}{27a2c3}

\definecolor{deepred}{HTML}{fe7b5b}

\usepackage[font=small,labelfont=bf]{caption}

\usepackage[font=footnotesize,labelfont=bf]{caption}

\definecolor{citecolor}{HTML}{faa700} 
\definecolor{lblue}{HTML}{ffb114} 
\definecolor{ogreen}{HTML}{2E7D32}
\definecolor{bred}{HTML}{BF360C}
\definecolor{newbrown}{HTML}{795548}

\hypersetup{
    colorlinks=true,
    linkcolor=orange,
    filecolor=magenta,      
    urlcolor=orange,
    citecolor=orange,
}

\begin{document}

\title{VB-Com: Learning Vision-Blind Composite Humanoid Locomotion Against Deficient Perception}

\author{\authorblockN{Junli Ren\textsuperscript{1,2} 
\quad Tao Huang\textsuperscript{1,3}
\quad Huayi Wang\textsuperscript{1,3}
\quad Zirui Wang\textsuperscript{1,4}
\quad Qingwei Ben\textsuperscript{1,5}\\
\quad Junfeng Long\textsuperscript{1}
\quad Yanchao Yang\textsuperscript{2}
\quad Jiangmiao Pang\textsuperscript{1, \dag}
\quad Ping Luo\textsuperscript{1,2, \dag}
}
\authorblockA{
\textsuperscript{1}Shanghai AI Laboratory \quad 
\textsuperscript{2}The University of Hong Kong \quad
\textsuperscript{3}Shanghai Jiao Tong University \quad \\
\textsuperscript{4}Zhejiang University \quad
\textsuperscript{5}The Chinese University of Hong Kong \\
Website: \href{https://renjunli99.github.io/vbcom.github.io/}{\texttt{vbcom.github.io}}
}
}

\twocolumn[{\renewcommand\twocolumn[1][]{#1}
\maketitle
\vspace{-0.5cm}
\begin{center}
    \centering
    \captionsetup{type=figure}
     \includegraphics[width=1.0\textwidth]{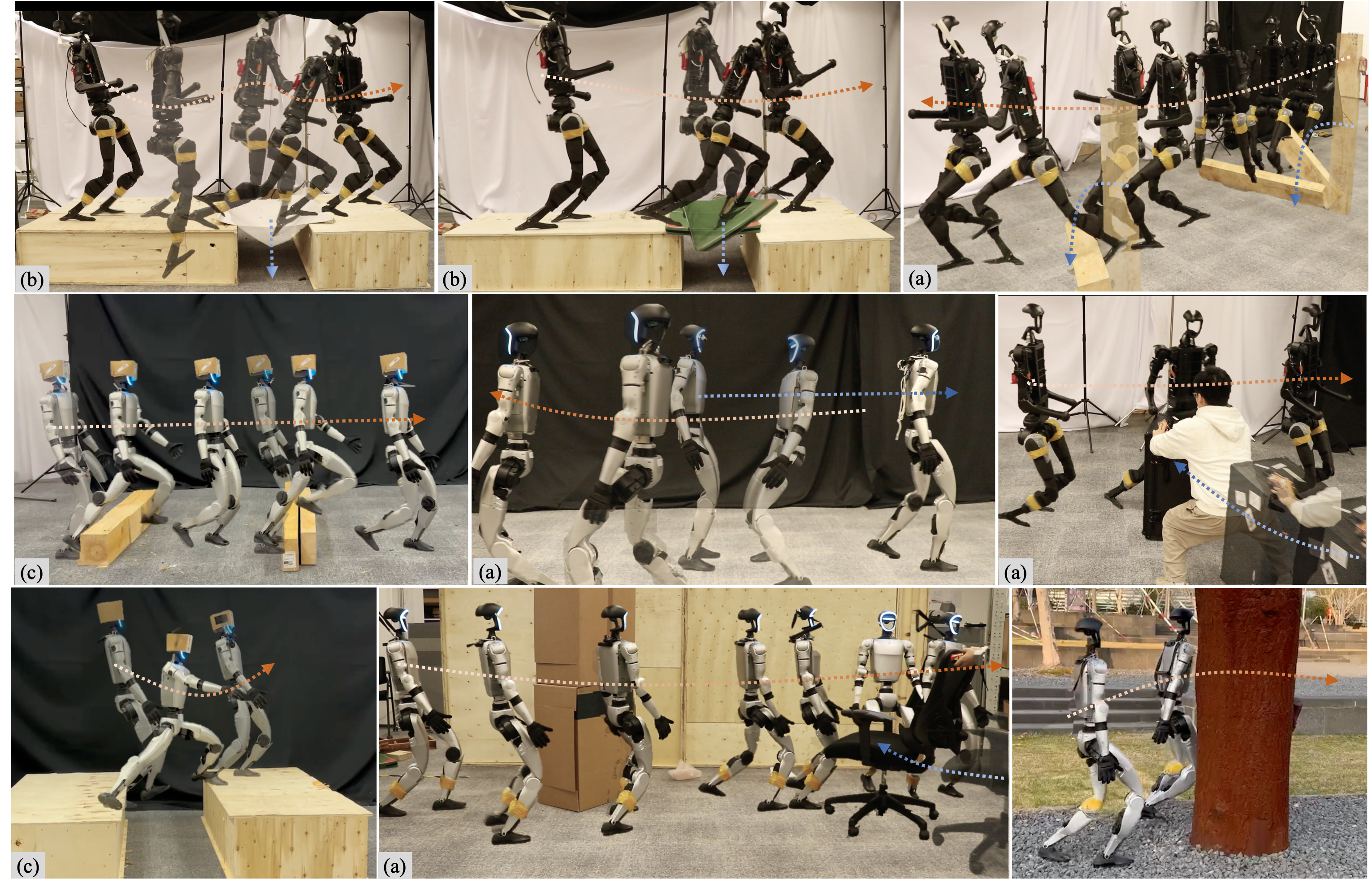}
     \vspace{-0.17in}
    \caption{\textbf{Overview.} VB-Com enables humanoid robots (\textcolor{orange}{move direction in orange arrorw}) to traverse dynamic terrains and obstacles (\textcolor{blue}{move direction in blue arrorw}). The perception deficiency is introduced by (a) suddenly appearing obstacles/hurdles, (b) deformable gaps, and (c) sensor occlusions. We demonstrate the effectiveness of VB-Com on both robots of Unitree G1 and H1.}
    \label{fig:teaser}
\end{center}
\vspace{-0.0in}
}]

\begin{abstract}
The performance of legged locomotion is closely tied to the accuracy and comprehensiveness of state observations. ``Blind policies", which rely solely on proprioception, are considered highly robust due to the reliability of proprioceptive observations. However, these policies significantly limit locomotion speed and often require collisions with the terrain to adapt. In contrast, ``Vision policies" allows the robot to plan motions in advance and respond proactively to unstructured terrains with an online perception module. However, perception is often compromised by noisy real-world environments, potential sensor failures, and the limitations of current simulations in presenting dynamic or deformable terrains. Humanoid robots, with high degrees of freedom and inherently unstable morphology, are particularly susceptible to misguidance from deficient perception, which can result in falls or termination on challenging dynamic terrains. To leverage the advantages of both vision and blind policies, we propose VB-Com, a composite framework that enables humanoid robots to determine when to rely on the vision policy and when to switch to the blind policy under perceptual deficiency. We demonstrate that VB-Com effectively enables humanoid robots to traverse challenging terrains and obstacles despite perception deficiencies caused by dynamic terrains or perceptual noise.
\end{abstract}      

\IEEEpeerreviewmaketitle

\section{Introduction} \label{sec:intro}

While legged locomotion control has been well-addressed through reinforcement learning with effective data collection \cite{makoviychuk2021isaac, gu2024humanoid, Genesis, zakka2025mujoco} and well-crafted reward guidance \cite{margolis2023walk, hwangbo2019learning, radosavovic2024learning, chen2024learning, li2023robust, long2024learning, margolis2024rapid}, the performance of such policies remains highly dependent on the accuracy and comprehensiveness of state observations \cite{nahrendra2023dreamwaq, sun2024leg, long2024hybrid, margolis2024rapid}. The state space can be roughly categorized into three types: 1) Accessible states, which are reliable and obtainable on real robots, such as joint encoders and IMU; 2) Privileged states \cite{kumar2021rma, lee2020learning}, which are unavailable on real robots, including velocity and static hardware parameters; and 3) External states \cite{miki2022learning, li2023robust, zhuang2024humanoid, long2024learning, yang2021learning}, which are observable but inherently noisy and occasionally unreliable. Previous works \cite{lee2020learning, kumar2021rma} have focused on encoding historical accessible states to approximate privileged and external states and these attempts on quadrupeds achieves successful traversal of static unstructured terrains such as stairs and slopes. However, these estimation methods often require robots to physically interact with unstructured terrains before responding \cite{long2024hybrid, cuiadapting, margolis2023walk}, forcing a trade-off between sacrificing speed to ensure safety or acting quickly but failing in scenarios that demand rapid responses, potentially leading to falls. 

To address this, perceptive locomotion methods have been developed \cite{cheng2024extreme, agarwal2023legged, yang2023neural}, enabling robots to anticipate incoming terrains and plan motions in advance using onboard sensors to obtain external states. Despite their impressive results, these methods are heavily dependent on maintaining consistency between perceived external states and those appeared during simulation \cite{hoeller2024anymal}. When mismatches occur, the robot may exhibit abnormal or dangerous behaviors.

In practice, it is impossible to provide the robot with all potentially encountered external states within the simulator \cite{zhu2025vr}. Current contact models in simulators are limited to rigid-body interactions and it is computationally expensive to incorporate dynamic terrains and obstacles during training \cite{choi2023learning}. Although previous research has highlighted the combination of perception and proprioception to achieve robust locomotion performance \cite{miki2022learning, zhang2024resilient, fu2022coupling} against perception inaccuracy, these studies have primarily focused on quadruped robots and low-risk scenarios, where a delayed response to the environment does not typically lead to locomotion failure.

Despite the impressive results of recent research achieving humanoid motions through tele-operation and imitation learning \cite{cheng2024expressive, ji2024exbody2, lu2024mobile, he2024hover, fu2024humanplus, he2024omnih2o}, the bipedal lower-limb structure of humanoid robots presents unique challenges in locomotion control compared to quadrupeds \cite{gu2024advancing, radosavovic2024learning}. The shifting of the gravity center in humanoid robots makes them more prone to unrecoverable falls. As a result, humanoid robots are more vulnerable to unexpected physical interactions given deficient perception. Consequently, current perceptive humanoid locomotion studies are limited to static terrains and confined environments, with performance heavily reliant on the quality of the perception module \cite{long2024learning, zhuang2024humanoid}.

In this work, we propose \textbf{VB-Com} (Vision-Blind Composite Humanoid Control), a locomotion policy capable of handling dynamic obstacles and compensating for deficient perception. VB-Com enables the robot to determine when to trust the perception module for accurate external state observations and when to disregard it to avoid misleading information that could result in locomotion failures. To achieve this, we first develop a vision policy that utilizes external visual observations from an onboard perception module, and a blind policy that relies solely on proprioceptive observations. The policies are combined using two return estimators, trained alongside the locomotion policies. These estimators predict future returns for each policy based on the current state and determining whether to rely on the vision policy or switch to the blind policy. As demonstrated in Fig \ref{fig:teaser}, in situations where the onboard sensors fail to provide comprehensive perception, VB-Com effectively enables the robot to quickly recover from potential failures caused by deficient perception, allowing it to traverse challenging terrains and obstacles. The contributions of this work can be summarized as follows:

\begin{itemize}
    \item A perceptive and a non-perceptive humanoid locomotion policy that can traverse gaps, hurdles and avoid obstacles.
    \item A novel hardware-deployable return estimator that predicts future returns achieved by current policy conditioned on proprioceptive states observation.
    \item A dual-policy composition system that integrates vision and blind policies for robust locomotion through dynamic obstacles and terrains where onboard sensors provide deficient external perception.
\end{itemize}

\section{Related Work} \label{sec:relatedwork}

\subsection{Robust Perceptive Legged Locomotion}

Typically, perceptive legged locomotion policies encode external state observations from onboard sensors as inputs of the policy network, allowing the robot to plan motions in advance to navigate unstructured terrains and avoid penalties for collisions or imbalance \cite{long2024learning, hoeller2024anymal, he2024agile}. Generally, lidar-based elevation maps \cite{miki2022elevation, hoeller2024anymal} and depth images \cite{cheng2024extreme, agarwal2023legged, yu2024walking, luo2024pie} are widely used to acquire external state observations. However, depth images are significantly affected by lighting conditions and limited field of view, while lidar-based elevation maps require time to construct, restricting their applicability to static environments.

Since comprehensive perception cannot be guaranteed on hardware, later studies have focused on integrating proprioceptive and exteroceptive observations to achieve robust locomotion or navigation against deficient perception \cite{miki2022learning, chen2024identifying, zhang2024resilient, fu2022coupling, ren2024top}. These approaches either employ a belief encoder that integrates exteroceptive and historical proprioceptive observations, or they address deficient perception at the path planning level. However, none of these methods have demonstrated the ability to perform rapid recovery actions in scenarios where deficient perception could quickly lead to failure, such as stepping into an unobserved gap and regaining balance before falling.

In this work, we address the mentioned challenge through policy composition: The blind policy is activated when deficient perception disrupts locomotion. Since both policies share the same state and action space, the proposed method allows the robot to recover from such situations quickly and safely.

\subsection{Hierarchical Reinforcement Learning}
Hierarchical reinforcement learning has been extensively explored in the literature, with the composition of low-level skills emerging as a popular approach for addressing long-horizon or complex tasks \cite{peng2019mcp, gupta2023bootstrapped, bacon2017option}. Among these works, value functions play a crucial role in policy composition \cite{shah2021value, nasiriany2022augmenting, zhang2023policy}, particularly in capturing the affordances of each sub-task. VB-Com draws inspiration from these approaches by training two return estimators, each representing the capabilities of the vision and blind policies, respectively.

In addition, several works in legged locomotion have explored hierarchical structures, such as employing DAgger to distill a set of locomotion skills \cite{zhuang2023robot}. Recent research \cite{he2024agile} also proposed a switching mechanism to achieve high-speed locomotion while avoiding obstacles. However, these frameworks rely heavily on vision observations, making them intolerant to perception outliers. In contrast, VB-Com addresses the novel challenge of maintaining stable locomotion despite deficient perception, with a specific focus on humanoid robots and high-dynamic tasks.
\section{Preliminaries} \label{sec:preliminaries}

\subsection{Probelm Formulation}
Reinforcement Learning based locomotion control is commonly modeled as a Partially Observable Markov Decision Process (POMDP), characterized by the tuple $(\mathcal{S}, \mathcal{A}, \mathcal{O}, \mathcal{R})$. In this formulation, the state space $\mathcal{S}$ represents the full state of the robot and environment, including privileged information and accurate terrain maps, whereas the observation space $\mathcal{O}$ encompasses only partial and noisy observations obtained from onboard sensors. 

The control policy $\pi(a|o)$, typically represented by a neural network, maps observations $o \in \mathcal{O}$ to actions $a \in \mathcal{A}$. Given the reward functions $r \in \mathcal{R}$ and a discount factor $\gamma$, the policy is trained to maximize the expected cumulative return:
\begin{equation}
    J(\pi) =  \mathbb{E}_{a_t \sim \pi(o_t)}[\sum_{t} ^ \infty \gamma^t r(s_t, a_t)].
\end{equation}

In this work, we address a more challenging POMDP task where the partial observations $\mathcal{O}$ include a potentially unreliable component $o_v$, which can fail under certain conditions, leading to significant penalties or even task termination. However, completely discarding $o_v$ would substantially limit the theoretical performance of the policy. The ideal solution is to enable the policy to recognize when $o_v$ becomes unreliable and switch to relying solely on the reliable proprioceptive observations $o_p$. We propose a composite solution to address this challenge in this work.

\subsection{Return Estimation}

Building upon previous research, we employ Proximal Policy Optimization (PPO) to train the locomotion policy for its effectiveness in continuous control. In addition to the actor policy $\pi(a|o)$, we emphasize that a well-trained policy also provides a practical approximation of the return $Q(s, a \sim \pi)$ during the Generalized Advantage Estimation (GAE) ~\cite{schulman2015high} process. The approximated composite returns will be implemented to develop the proposed policy composition mechanism.

\subsection{Q-informed Policies Composition}

Given a set of policies $\Pi = \{\pi_1, \pi_2, \dots, \pi_n \}$ that share the same states, actions, and rewards $(\mathcal{S}, \mathcal{A}, \mathcal{R})$, a composite policy $\tilde{\pi}$ selects an action from the proposed action set $A = \{ a_i \sim \pi_i(s) \}$ with a probability $P_w$ that is related to their respective potential utilities. In the context of Markov Decision Process, it has been proved \cite{zhang2023policy} that selecting actions based on the cumulative return at current states and candidate actions will achieve the best expected return for $\tilde{\pi}$. To this end, the Q-value based policies composition will compute the cumulative return at current for each low-level policy $\bold{Q} = \{ Q_i(s, a_i) | a_i \in \mathcal{A} \}$ and construct a categorical distribution to select the final action:

\begin{equation}
    P_w(i)= \frac{exp(Q_i(s, a_i)/\alpha)}{\sum_j exp(Q_j(s, a_j)/\alpha)},
\end{equation}
here $\alpha$ is the temperature. In the case of two sub-policies, such composition will assign a higher probability to the action with the higher Q-value at the current state.

\section{Method} \label{sec:method}

\begin{figure*}[h]
\centering{\includegraphics[width=\textwidth]{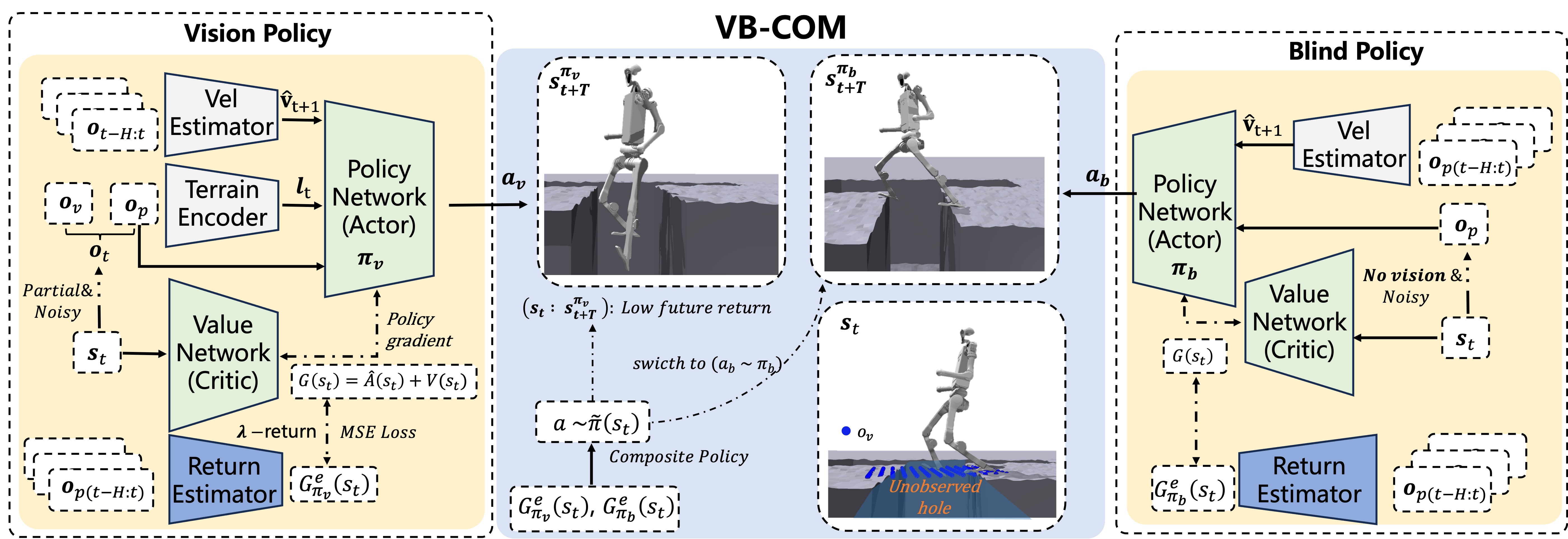}}
\caption{Overview of our framework: In VB-Com, we develop two locomotion policies—one perceptive and one non-perceptive—through single-stage training. These sub-policies are integrated based on two return estimators, which predict future returns given the current state for each of the policy policy. This integration enables seamless policy switching, allowing the robot to effectively adapt to varying levels of perceptual information and handle dynamic environments.}
\label{method}
\end{figure*}

\subsection{System Overview}

The proposed VB-Com framework (Fig \ref{method}) comprises a perceptive locomotion policy $\pi_v$ and a non-perceptive policy $\pi_b$. $\pi_v$ incorporates visual observations to enable perceptive locomotion, allowing humanoid robots to traverse complex terrains such as high steps, gaps, hurdles, and to perform obstacle avoidance. $\pi_b$ is trained within the same reward and action space but does not accept external observations.

Once well-trained, $\pi_v$ and $\pi_b$ are expected to operate stably on familiar terrains with corresponding observations provided during training. Under such conditions, the VB-Com framework prioritizes selecting actions from the vision-based policy $\pi_v$ due to its more comprehensive observation of the environment and higher expected return. However, when the robot encounters outlier scenarios, such as perceptive deficiencies that the environments do not interact with the robot as predicted by the vision-based observations $o_v$, the blind policy $\pi_b$ takes over, leveraging the relatively reliable proprioceptive observations $o_p$ to navigate such situations effectively.

VB-Com achieves the mentioned composition with two return estimators, $\pi_v^e$ and $\pi_b^e$, trained concurrently with the locomotion policies. The estimators provide approximations of the cumulative return that the system will obtain whther chooses the vision or blind policy at the current step. During deployment, the compositor operates based on the estimations of the returns $\{\hat{G}^e_v \sim \pi_v^e, \hat{G}^e_b \sim \pi_b^e \}$, from which one executed action is selected from the candidate actions $\{a_v \sim \pi_v, a_b \sim \pi_b \}$.
 
\subsection{Locomotion Policies}

To demonstrate the quick responsiveness of VB-Com in handling deficient perception, we train the locomotion policies on challenging terrains, including gaps, hurdles, and high walls (for obstacle avoidance), which require high joint velocities for traversal. This contrasts with the more common scenarios, such as stairs and discrete steps, which have been the focus of prior works. Drawing from previous experience, we adopt a goal-reaching formulation rather than velocity tracking to train the policies, as this approach is better suited for completing highly dynamic tasks.

\subsubsection{Observation Space}

The policy observations $o_t$ consist of two components: $o_p$, which includes the commands and proprioceptive observations, and $o_v$, which represents the visual observations.

The commands are designed following \cite{cheng2024extreme}, where directions are computed using waypoints placed on the terrain: $\textbf{d} = (\textbf{p} - \textbf{x}) / ||\textbf{p} - \textbf{x}||$, with $\textbf{p}$ and $\textbf{x}$ representing the locations of the waypoints and the robot, respectively. To prevent sharp directional changes, the robot is provided with directions to the next two goal locations at each step, along with linear velocity commands $\textbf{v}_c$. These commands are represented as a three-dimensional vector: $\textbf{c}_t = [\textbf{d}_1, \textbf{d}_2, \textbf{v}_c]$. The proprioceptive observations consists of its joint position $\mathbf{\theta}_t$, joint velocity $\dot{\mathbf{\theta}}_t$, base angular velocity $\mathbf{\omega}_t$ and gravity direction in robot frame $\mathbf{g}_t$. The perceptive information $o_v$ is a robotic-centric height map around the robot, as the hardware implementation detailed in ~\cite{long2024learning}. The perceptive observation is not provided while training the blind policy $\pi_b$.

As stated in the problem formulation, $o_t$ represents the observation space of the POMDP. Therefore, both $o_p$ and $o_v$ are domain-randomized during training to better simulate sensor noise encountered in real-world scenarios. On the other hand, the critic network, which is responsible for providing the actor policy with accurate state evaluations, is allowed to access privileged information. Building upon previous research, we incorporate the accurate linear velocity $v_t$, which plays an essential role in legged locomotion tasks, as the additional privileged information. Meanwhile, the proprioceptive and perceptive states used in the critic network are not subjected to noise. Additionally, we provide a larger height map to the critic network ($1.6m \times 1.0m$) compared to the one used in the actor network ($1.2m \times 0.7m$), which we found facilitates faster adaptation of the robot to higher terrain levels during curriculum learning.

\subsubsection{Rewards}

The majority of our reward functions are adapted from \cite{long2024learning, cheng2024extreme}. To align with the goal-reaching commands in the observation space, we modify the task reward to track the target direction and linear velocity commands instead of angular velocity. We also include a series of regularization rewards to encourage the humanoid robot to exhibit natural locomotion and maintain gentle contact with the ground. In addition, unlike previous research that treats obstacle avoidance as a path planning problem, we enable the robot to autonomously reach its goal while avoiding obstacles at the locomotion level. This is achieved through a carefully balanced trade-off between goal-reaching rewards and collision penalties. To acquire an accurate return estimation, we focus the rewards on the proprioceptive states of the robot rather than interactions with the environment. The reward scales for both the blind and vision policies remain consistent throughout the learning process. We achieve a unified policy capable of simultaneously traversing obstacles, hurdles, and gaps through the proposed reward setting.

\subsubsection{State Estimation}
A variety of approaches have been proposed in legged locomotion to address POMDPs by constructing a belief state from historical observations, often involving a second-stage training process or complex network structure. In this work, we propose an efficient and simple state estimator that predicts the next velocity $v_{t+1}$ based on the historical observation sequence $o_{t-H:t}$. Both $v_{t+1}$ and $o_{t-H:t}$ are rolled out online from the collected trajectories while the policy is being updated. A regression loss is used to update the state estimator. We demonstrate that, with the state estimator, a hardware-deployable locomotion policy can be achieved through a single stage of training, enabling agile locomotion tasks with high effectiveness and data efficiency.

\subsection{Vision-Blind Composition}
\label{subsec:vb-com}
Given the vision policy $\pi_v$ and the blind policy $\pi_b$, the composition can be viewed as a discrete policy $\tilde{\pi}$ with an action dimension of two, selecting between the candidate actions:

\begin{equation}
    \tilde{\pi}(a|s) = [a_b \sim \pi_b, a_v \sim \pi_v ]\textbf{w}, \textbf{w} \sim P_\textbf{w},
\end{equation}
Building on the analysis of Q-informed policy composition, for each state $s_t$ at each step, we have:
\begin{equation}
    P_\textbf{w}(i|s_t,a_v,a_b) \propto exp(Q(s_t,a_i)), a_i \in \{a_v,a_b\} \label{propto}.
\end{equation}

\subsubsection{Policy Return Estimation}
Given the current states $s_t$ of the robot, we can estimate the expected cumulative return $G_{\pi_i}(s_t)$ for each policy to guide the composition process. In practice, switching between the two independently trained policies could cause abrupt changes in the performed actions. For example, a switch from the vision policy to the blind policy should help the robot avoid falling into an unseen gap, which may require a sequence of actions from the blind policy without the involvement of vision. To address this, we introduce a switch period $T$, which acts as the control unit for each switch. The introduction of $T$ also helps decouple the switching actions, approximately making them temporally independent of each other.

To this end, we expect the return estimator to be responsible for estimating a time sequence of expected returns over the duration $T$, such that:
\begin{equation}
    L_{\pi_i} = \mathbb{E}_t[\hat{G}_{\pi_i}^e(s_t) - G_{\pi_i}(s_{t:{t+T}})],
\end{equation}
To achieve the estimation with reduced bias and variance, we implement $\lambda$-return to weight the time-sequenced returns within one switch period as follows:
\begin{equation}
    G_{\pi_i}(s_{t:{t+T}}) \approx G^\lambda_{\pi_i}(s_{t:{t+T}}) = (1-\lambda)\sum_{n=t}^{t+T}\lambda^{n-1}G_{\pi_i}(s_{n}),
\end{equation}
which represents a weighted return if the robot chooses to switch to the low-level policy $\pi_i$ given the state $s_t$. In addition, in order to mitigate the large variance between single-step rewards and prevents the policy from overfitting to recent batches, $G_{\pi_i}$ is computed based on the update of value functions ~\cite{schulman2015high}, where $G_{\pi_i}(s_t) = \hat{A}(s_t) + V(s_t)$, with $\hat{A}(s_t)$ being the advantage function and $V(s_t)$ the value function.

Since the return estimators need to be deployable on hardware and we aim to mitigate perception misleadings, we avoid using exteroceptive observations or privileged information as inputs. Instead, we use the historical proprioceptive observation sequence $o_{p_{t-H:t}}$ as the input to the return estimator $\pi^e$.

\subsubsection{Policy Switch}
Unlike previous works that construct a switch-based hierarchical framework to keep the robot within a safe domain and prevent potential collisions, VB-Com performs policy switching to recover the robot from getting stuck due to perceptive deficiencies.

Ideally, equation \ref{propto} provides the theoretical basis for choosing the action with the greater value estimation $\hat{G}_\pi^e$ at the current state. This aligns with the fact that $\pi_v$ typically yields higher returns than $\pi_b$ as long as the vision observations are consistent with those seen during training, since $\pi_v$ has access to more comprehensive environmental observations.

During deployment, when the robot experiences a sudden environmental change that disrupts locomotion, both estimations $\hat{G}_{\pi_{v,b}}^e$ will decline. We observe that in these situations, it is difficult to maintain strict monotonicity such that $\hat{G}^e_{\pi_b} > \hat{G}^e_{\pi_v}$ due to the return approximation error introduced by $\pi^e$. Meanwhile, the blind policy demonstrates greater sensitivity to unstable motions, as the low-return samples are more frequently encountered even after the policy has been well-trained, compared to $\pi_v$ (as illustrated in Fig. \ref{return}). To address this, we introduce a threshold $G_{th}$ trigger that can also prompt the policy switch.
\begin{equation}
a \sim \tilde{\pi}(s_t) = \left\{
\begin{array}{ll}
a_v, & \text{if } G^e_{\pi_v}(s_t) > G^e_{\pi_b}(s_t)> G_{th} , \\
a_b, & \text{otherwise,}
\end{array}
\right.
\end{equation}
\begin{equation}
G_{th} =  1/5\sum_{t-5}^{t}{G_{\pi_b}^e(s_i)} - \alpha, 
\end{equation}
here $\alpha$ is a threshold hyperparameter. In practice, we replace $G^e_{\pi_v}(s_t)$ with a smoothed window (length 5) to avoid sudden abnormal estimations, which we have found to be effective in real robot deployments. Additionally, a switch will not be performed under conditions of high joint velocity to prevent potential dangers caused by the abrupt switching of policies when the robot is performing vigorous motion.

\begin{figure}[htbp]
\centering{\includegraphics[width=0.5\textwidth]{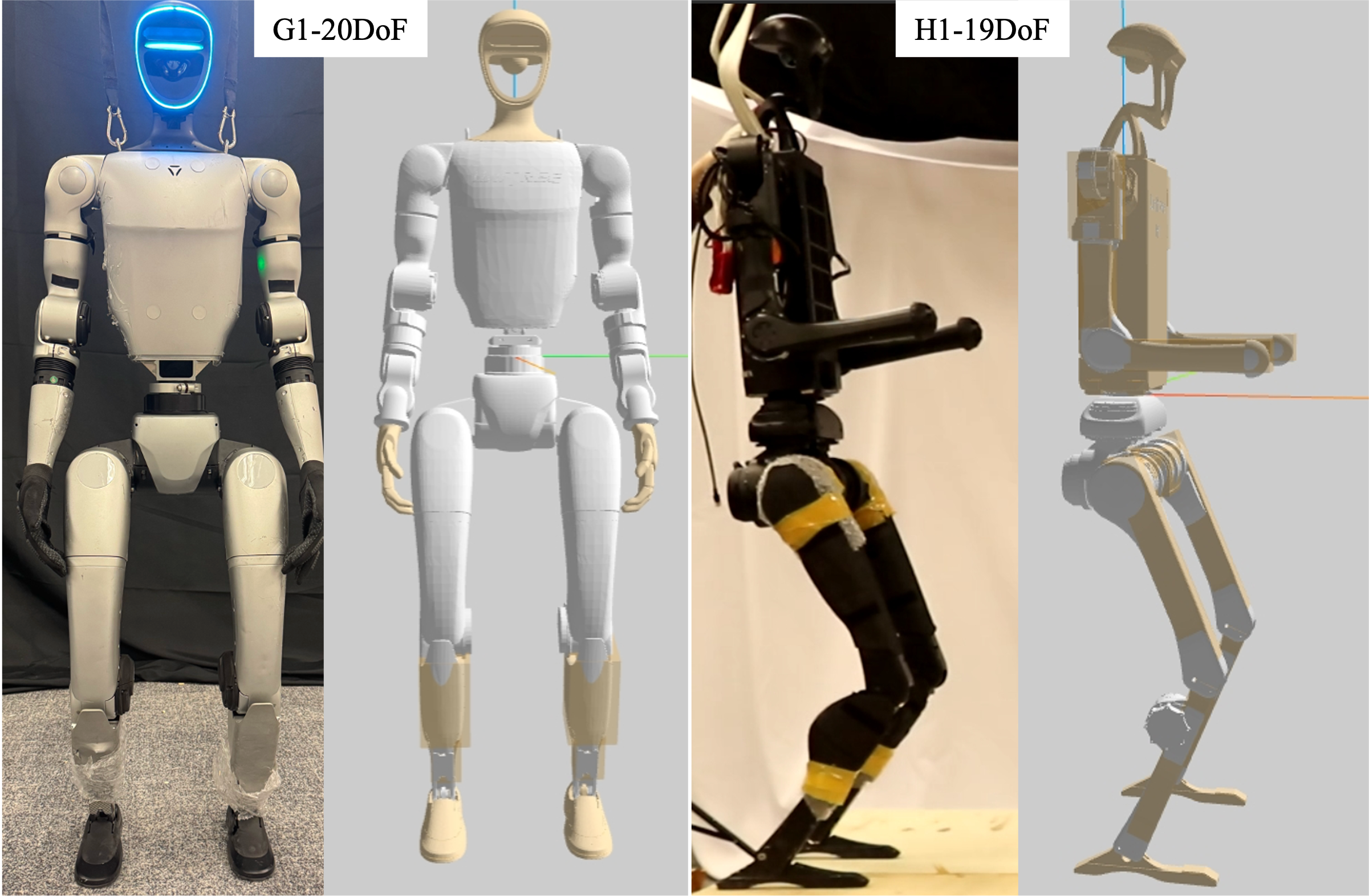}}
\caption{We train the proposed framework on Unitree G1 and H1 humanoid robots with the \textcolor{olive}{enabled collisions links $(C_l^e)$}.}
\label{robot}
\end{figure}
\subsection{Implementation Details}

\subsubsection{Humanoid Robots}
We implement VB-Com on two humanoid robots, Unitree-G1 (G1) and Unitree-H1 (H1), in both simulation and real-world environments (Fig \ref{robot}). Both robots are controlled using whole-body actions, with G1 having 20 control actions (10 upper-body and 10 lower-body joints) and H1 having 19 control actions (8 upper-body, 10 lower-body joints, and 1 torso joint). Since enabling all collision links for the robot can result in significant computational overhead (especially for G1), we activate a subset of collision links $(C_l^e)$ sufficient to accomplish the locomotion tasks, particularly for the blind policy where prior contact is necessary. For example, by enabling the hand collision on G1, the robot learns to reach out its hands to touch potential obstacles and avoid them once perception becomes deficient.

\subsubsection{Perception \& noise}
We implement a robotic-centric elevation map on both G1 and H1 to acquire external state observations for the vision policy. The lidars mounted on the robots' heads serve as onboard sensors. Since the elevation map requires a smooth time window to integrate newly observed point clouds into the map, it struggles with dynamic scenes, presenting vision-deficient challenges that can be effectively addressed by VB-Com.

We also introduce standard noise during the training of $\pi_b$ to enhance its tolerance against deficient perception (Training Noise in Fig \ref{noise}). This includes $10\%$ Gaussian noise and random perception delays within the past $0.5$ seconds. The added perception noise aims to achieve relatively deployable performance for $\pi_v$. However, we demonstrate that $\pi_v$ fails to handle a wider range of perception noise or dynamic obstacles encountered in real-world scenarios.

\section{Result} \label{sec:result}

\subsection{Setup}

In this section, we evaluate VB-Com across the following perspectives:
\begin{itemize}
    \item Under what conditions does VB-Com demonstrate superior performance compared to using a single-policy approach?
    \item How does VB-Com outperforms baseline methods in those scenarios?
    \item How well does the proposed return estimator contribute to the composition system?
\end{itemize}

\begin{figure}[h]
\centering{\includegraphics[width=0.5\textwidth]{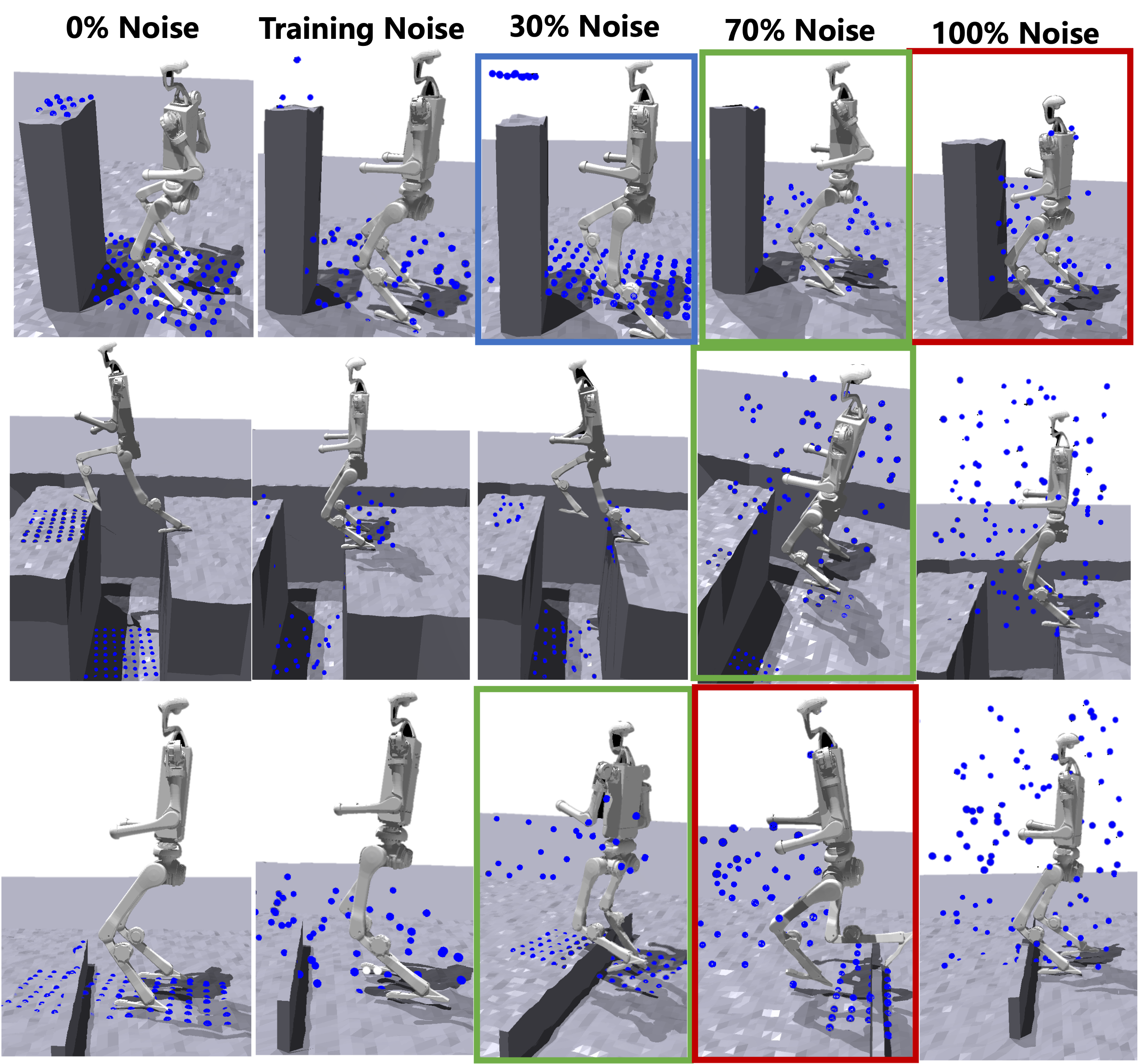}}
\caption{We present four types of perception noises and implement them on heightmaps during evaluation: gaussian noise, \textcolor{red}{forward shifting noise}, \textcolor{green}{lateral shifting noise} and \textcolor{blue}{floating noise}.}
\label{noise}
\end{figure}

\subsubsection{Evaluation Noise}
To simulate situations where the robot encounters perception outliers not present in the simulation, we introduce a quantitative curriculum noise designed to mimic varying levels of perception deficiency. As shown in Fig. \ref{noise}, we focus on four types of noise: (1) \textbf{Gaussian noise}: noise points sampled from a Gaussian distribution, to the original heightmap. The noise level is scaled from 0.0 to 1.0, where the training noise level corresponds to a 0.1 noise level in this scenario. (2) \textbf{Shifting noise}: replacing points in the original heightmap with noise sampled from a Gaussian distribution. The range of replacement points is controlled by the noise level, where a $100\%$ noise level results in a fully noisy heightmap. The shifting direction can either be along the heading direction (red box) or sideways (green box). (3) \textbf{Floating noise}: The heightmap is displaced vertically, either upwards or downwards, the floating noise simulates variations in terrain height. (blue box).

\begin{table}[!ht]
\caption{Terrain Size Scales (m)}
\label{tab:terrains}
\begin{center}
\renewcommand\arraystretch{1.25}
\begin{tabular}{lcccc}
\toprule[1.0pt]
Terrain & Length & Width & Heights\\
\midrule[0.8pt]

Gaps        & $(0.6, 1.2)$ & $(\bm{0.6}, \bm{0.8})$ & $(-1.8, -1.5)$\\  
Hurdles     & $(0.8, 1.0)$ & $(0.1, 0.2)$ & $(\bm{0.2}, \bm{0.4})$\\  
Obstacles   & $(\bm{0.2}, \bm{0.4})$ & $(0.2, 0.4)$ & $(1.4,1.8)$\\  

\bottomrule[1.0pt]
\end{tabular}
\end{center}
\end{table}

\subsubsection{Experiments Setup}
In simulation, we conduct $10 \times 3$ experiments for each method across three types of terrain, replicating the experiments three times to calculate the variance. Each episode involves the robot navigating through 8 goal points, with each goal paired with a corresponding challenging terrain or obstacle. The size of the terrains is set to the maximum curriculum terrain level, as shown in Table \ref{tab:terrains}. The bolded values indicate the primary factors that contribute to the difficulty for the terrain.

\subsubsection{Baselines}
We primarily compare VB-Com with the vision and blind policies operating independently. Additionally, as previous works have shown that robust perceptive locomotion can be learned by incorporating various perception noises during training \cite{miki2022learning}, we add a \textbf{Noisy Perceptive policy baseline} trained using the same noises implemented in the evaluation. This allows us to examine how well the proposed VB-Com policy performs compared to policies that have already seen the evaluation noises. The evaluation noises are introduced to the Noisy Perceptive policy in a curriculum format during training, which evolves with the terrain level.

\begin{figure*}[h]
\centering{\includegraphics[width=\textwidth]{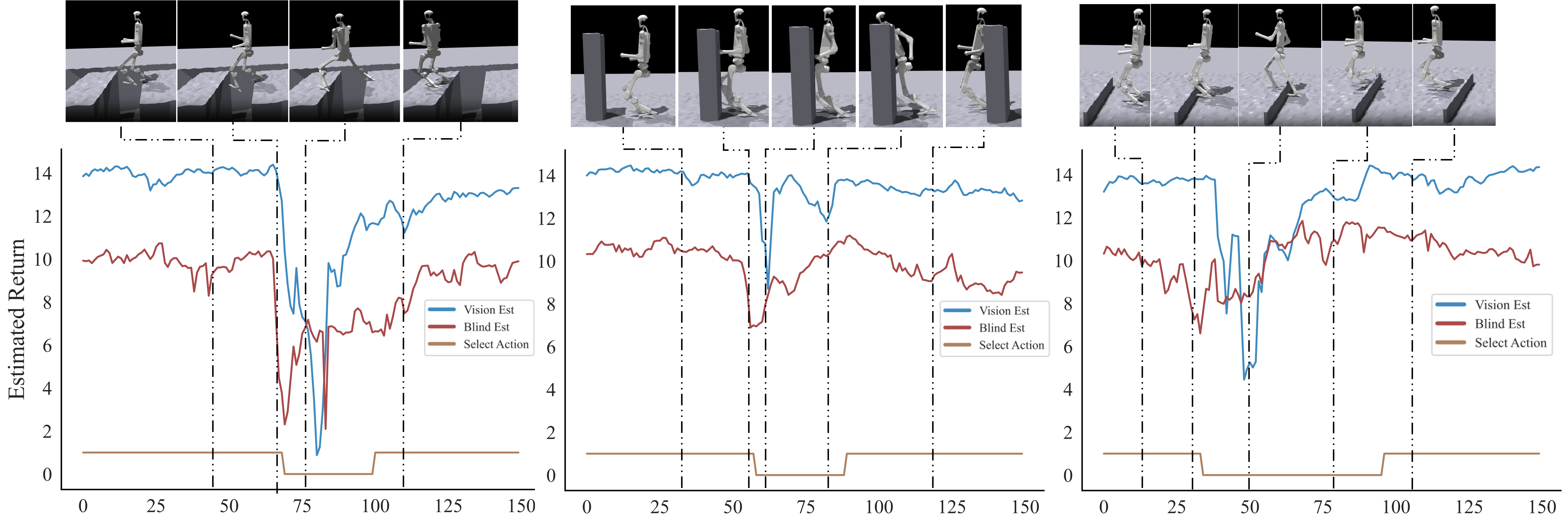}}
\caption{Illustrations of the variation in estimated return and action phases(0 for $a_b$ and 1 for $a_v$) across three concerned terrains.}
\label{return}
\end{figure*}

\subsection{Example Case}
First, we illustrate how VB-Com operates, specifically when the composition switches to $\pi_b$ and how it effectively controls the robot to traverse the terrain against deficient perception (Fig. \ref{return}). We demonstrate $3$ seconds of the estimated returns, along with the policy composition phase, as the robot walking through the challenging terrain during the simulation experiments at the noise level of $100\%$. Before the robot encounters challenging terrains, we observe that the estimated return $G^e_{\pi_v}(s_t)$ consistently exceeds $G^e_{\pi_b}(s_t)$, as the robot is walking on flat ground with relatively stable motion. This observation aligns with the discussion in Section \ref{subsec:vb-com}, where it was explained that $\pi_v$ benefits from the external state observation and results in a higher return $G_t$. This characteraistic ensures the robot operates at $\pi_b$ while stable motion. 

Once the deficient perception reaches the $100\%$ noise level, the robot will not be aware of the incoming challenging terrains until it collides with them. At this point, we observe that $G^e_{\pi}(s_t)$ drops sharply within several control steps, prompting the switch to the blind policy. This switch allows the robot to respond to the terrain, and once the motion stabilizes, $G^e_{\pi}(s_t)$ returns to a normal level, at which point the vision policy regains control. These cases demonstrate the effectiveness of VB-Com, which responds quickly to deficient perception, but avoids unnecessary switches to the blind policy when it is not needed.

\begin{table*}[!h]
\caption{VB-Com Evaluations}
\label{tab:VB-Com}
\begin{center}
\renewcommand\arraystretch{1.25}
\begin{tabular}{lccccccc}
\toprule[1.0pt]
Noise Level &Method & Goals Completed($\%$) & Rewards & Average Velocity & Fail Rate & Collision Steps($\%$) & Reach Steps\\
\midrule[0.8pt]


\multirow{2}{*}{0\% noise} & VB-Com & $84.05 \pm 2.28$ & \bm{$142.07 \pm 4.19$} & $0.71 \pm 0.01$ & \bm{$0.29 \pm 0.01$} & $1.50 \pm 0.14$ & $177.29 \pm 4.66$\\  
                              & Vision & $73.57 \pm 4.97$ & $118.07 \pm 10.42$ & $0.73 \pm 0.01$ & $0.42 \pm 0.07$ & \bm{$1.39 \pm 0.53$} & $204.82 \pm 28.91$\\  \midrule
\multirow{2}{*}{30\% noise} & VB-Com & $82.24 \pm 6.6$ & $132.81 \pm 7.64$ & $0.71 \pm 0.01$ & $0.34 \pm 0.10$ & $2.09 \pm 0.13$ & $178.13 \pm 4.13$\\  
                              & Vision & $72.76 \pm 2.29$ & $115.20 \pm 2.43$ & $0.75 \pm 0.02$ & $0.43 \pm 0.05$ & $2.52 \pm 0.32$ & $195.58 \pm 21.98$\\  \midrule
\multirow{2}{*}{70\% noise} & VB-Com & $82.48 \pm 1.20$ & $132.44 \pm 6.17$ & $0.70 \pm 0.02$ & $0.33 \pm 0.03$ & $2.12 \pm 0.11$ & $184.81 \pm 4.47$\\  
                              & Vision & $55.38 \pm 3.33$ & $58.24 \pm 13.97$ & $0.73 \pm 0.03$ & $0.67 \pm 0.07$ & $6.08 \pm 0.82$ & $190.50 \pm 18.28$\\  \midrule
\multirow{3}{*}{100\% noise} & VB-Com & \bm{$84.81 \pm 6.45$} & $129.99 \pm 9.84$ & $0.72 \pm 0.02$ & \bm{$0.29 \pm 0.08$} & $2.60 \pm 0.68$ & $182.29 \pm 11.47$\\  
                              & Vision & $48.71 \pm 5.60$ & $47.53 \pm 17.55$ & $0.70 \pm 0.06$ & $0.69 \pm 0.06$ & $6.92 \pm 1.36$ & $268.40 \pm 57.11$\\  
                              & Noisy Perceptive & $80.52 \pm 0.91$ & $116.94 \pm 4.07$ & \bm{$0.76 \pm 0.02$} & $0.32 \pm 0.04$ & $3.49 \pm 0.38$ & \bm{$154.98 \pm 4.41$}\\ \midrule
& Blind & $83.76 \pm 1.35$ & $131.29 \pm 3.48$ & $0.70 \pm 0.01$ & $0.33 \pm 0.05$ & $2.57 \pm 0.27$ & $184.08 \pm 1.85$\\  


\bottomrule[1.0pt]
\end{tabular}
\end{center}
\end{table*}

\subsection{Evaluations on Different Noise Levels}
\textbf{VB-Com achieves robust locomotion performance under different levels of perception deficiency.} As shown in Tab \ref{tab:VB-Com}, performance of the vision policy declines shaprly with the arise of noise level. In addition, since the evaluation experiments set the terrain curriculum to the maximum level, the vision policy struggles even at a $0\%$ noise level: It only achieves around $73\%$ goal-reaching success, with a termination rate exceeding $40\%$. This poor performance is likely due to the severe challenge terrains, such as the farthest range of the heightmap $(0.85m)$ is only $0.05m$ wider than the width of the gaps$(0.8m)$. In contrast, VB-Com achieves a stable higher goal-reaching success against different levels of perception deficiency. In contrast, VB-Com achieves consistently higher goal-reaching success across varying levels of perception deficiency, including both noise and perception range limitations.

Despite the high goal-reaching success, we also include additional metrics to further analyze the performance. The reward values recorded throughout each episode indicate the proposed method’s ability to achieve both goal completion and collision avoidance. These rewards strongly correlate with the robot’s success in reaching the target while minimizing collisions. For instance, VB-Com at the $0\%$ noise level achieves the highest rewards$(142.07)$, although the goal completion rate$(84.05)$ is slightly lower compared to the trail in $100\%$ noise level $(84.81)$. This is because VB-Com switches to the blind policy more often in  $100\%$  noise level, resulting in more frequent collisions and lower rewards obtained. 

The reach steps metrics indicates the smoothness of the policy in overcoming challenging obstacles. Since the switching mechanism requires several steps to respond effectively, VB-Com results in a higher number of reach steps as the noise level increases. This is because, under higher noise conditions, the system needs additional time to transition from the vision policy to the blind policy, which leads to more gradual and controlled responses to terrain challenges.
\begin{figure}[h]
\centering{\includegraphics[width=0.5\textwidth]{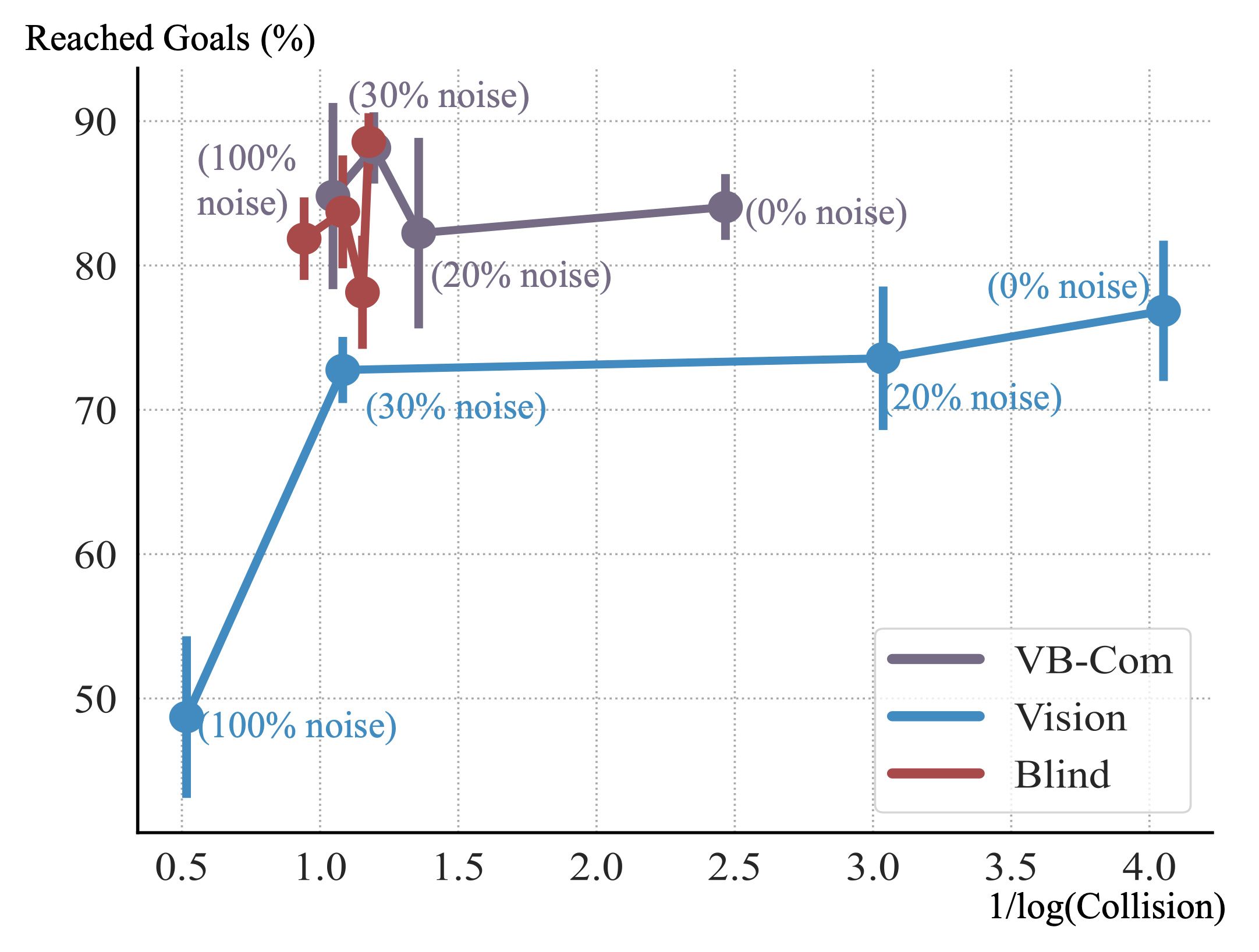}}
\caption{We compare the collision and goal-reaching performances under different noise levels. VB-Com achieves low collisions and high success rates with accurate perception, and its success rate remains high under deficient perception.}
\label{noiseevalueate}
\end{figure}

\begin{figure}[h]
\centering{\includegraphics[width=0.5\textwidth]{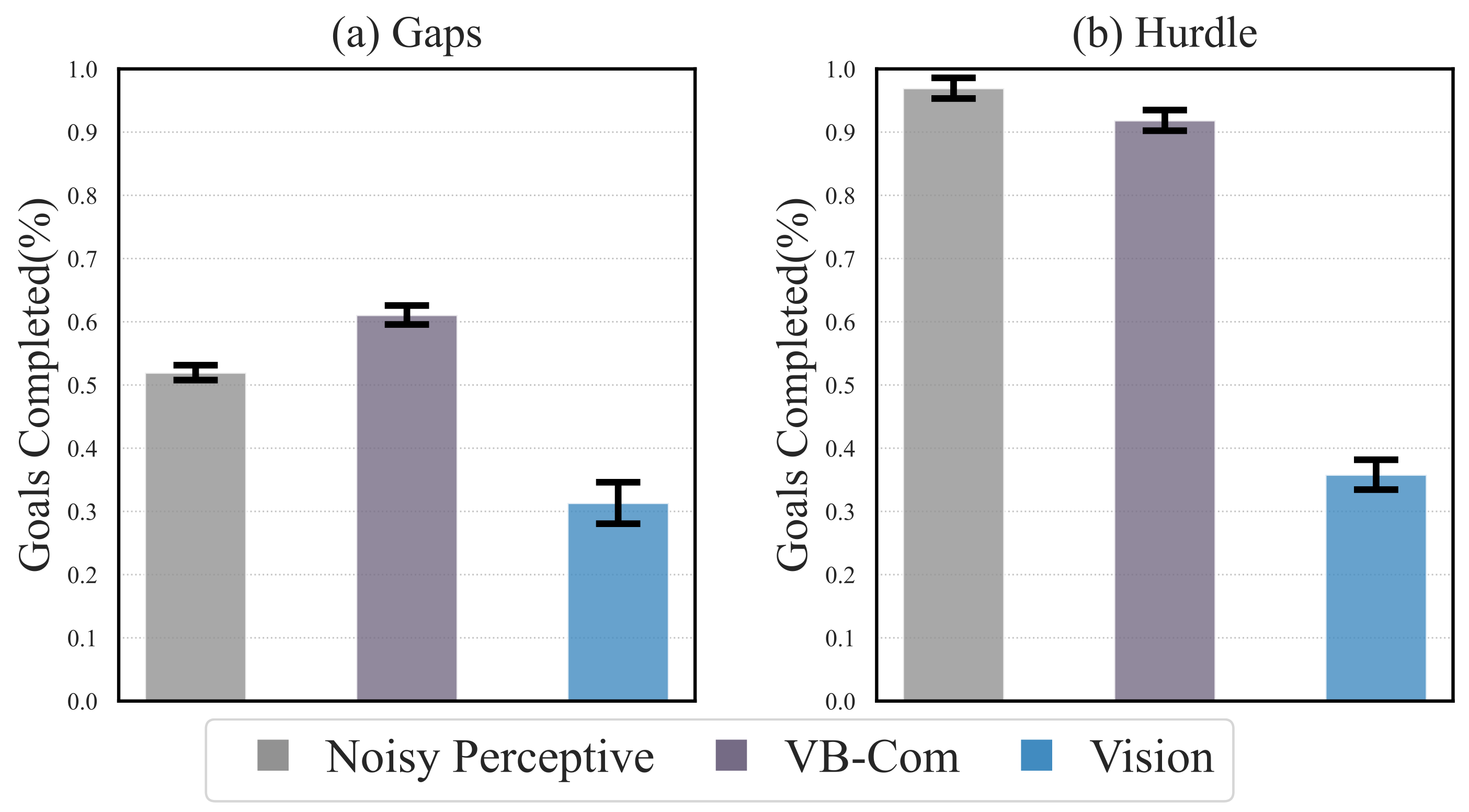}}
\caption{Comparisons between the Noisy Perceptive policy and VB-Com in navigating gaps and hurdles separately.}
\label{terraineval}
\end{figure}

\subsection{Comparisons with Blind Policy}
\textbf{VB-Com achieves less collision than the blind policy when perception becomes less dificient.} As shown in Tab \ref{tab:VB-Com}, the blind policy achieves a relatively high Goals Completed rate $(83.76\%)$, as its performance is unaffected by deficient perception. Therefore, we include an evaluation of the collision performance between VB-Com and the blind policy to further highlight the advantage of the proposed framework. In our evaluations, "Collision Steps" is defined as the ratio of the number of steps during which the robot collision model (Fig \ref{robot}) makes illegal contact with the terrain or obstacles, relative to the total number of steps within an episode.

We can observe from Tab \ref{tab:VB-Com} that the collision steps increase with the noise level for VB-Com. Fig \ref{noiseevalueate} provides a more intuitive illustration: as perception becomes more comprehensive, VB-Com achieves both fewer collisions and better goal-reaching performance. In contrast, the blind policy maintains a high goal-reaching rate but results in more collisions, while the vision policy performs better in avoiding collisions when the perception is accurate and comprehensive. As the noise level increases, the performance of VB-Com begins to resemble that of the blind policy. These results demonstrate the effectiveness of the composition system, which benefits from both sub-policies to achieve better performance in terms of both goal-reaching and minimizing collisions.

\subsection{Comparisons with Noisy Perceptive Training}
\textbf{Compared to policies trained with noisy priors, VB-Com achieves equivalent performance without prior knowledge of the noise, while also demonstrating better training efficiency and the ability to handle more challenging terrain difficulties.} The comparisons (Tab \ref{tab:VB-Com}) with Noisy Perceptive policy show that the Noisy Perceptive policy achieves a relatively high goal completion rate $(80.52\%)$ but exhibits a higher collision step rate $(3.49\%)$. It can be concluded that, as severe noise is introduced during evaluation, the heightmap quickly becomes random noise with the increasing noise level. In response, the Noisy Perceptive policy begins to exhibit behavior similar to that of the blind policy—making contact with obstacles and reacting when the noisy signals overwhelm the external observations.

To further investigate the conditions under which the Noisy Perceptive policy fails to surpass the performance of VB-Com, we evaluate goal-reaching performance under different terrains (Fig. \ref{terraineval}). The results show that VB-Com outperforms the Noisy Perceptive policy in gap terrains, while the Noisy Perceptive policy performs better in hurdle situations, achieving a higher success rate in preventing the robot from being tripped by hurdles. However, recovering from missed gaps requires a quicker response, or the robot risks falling. These results demonstrate that the single-policy method fails to handle such dynamic challenges effectively, highlighting the advantages of the composition in VB-Com.

\begin{figure}[h]
\centering{\includegraphics[width=0.5\textwidth]{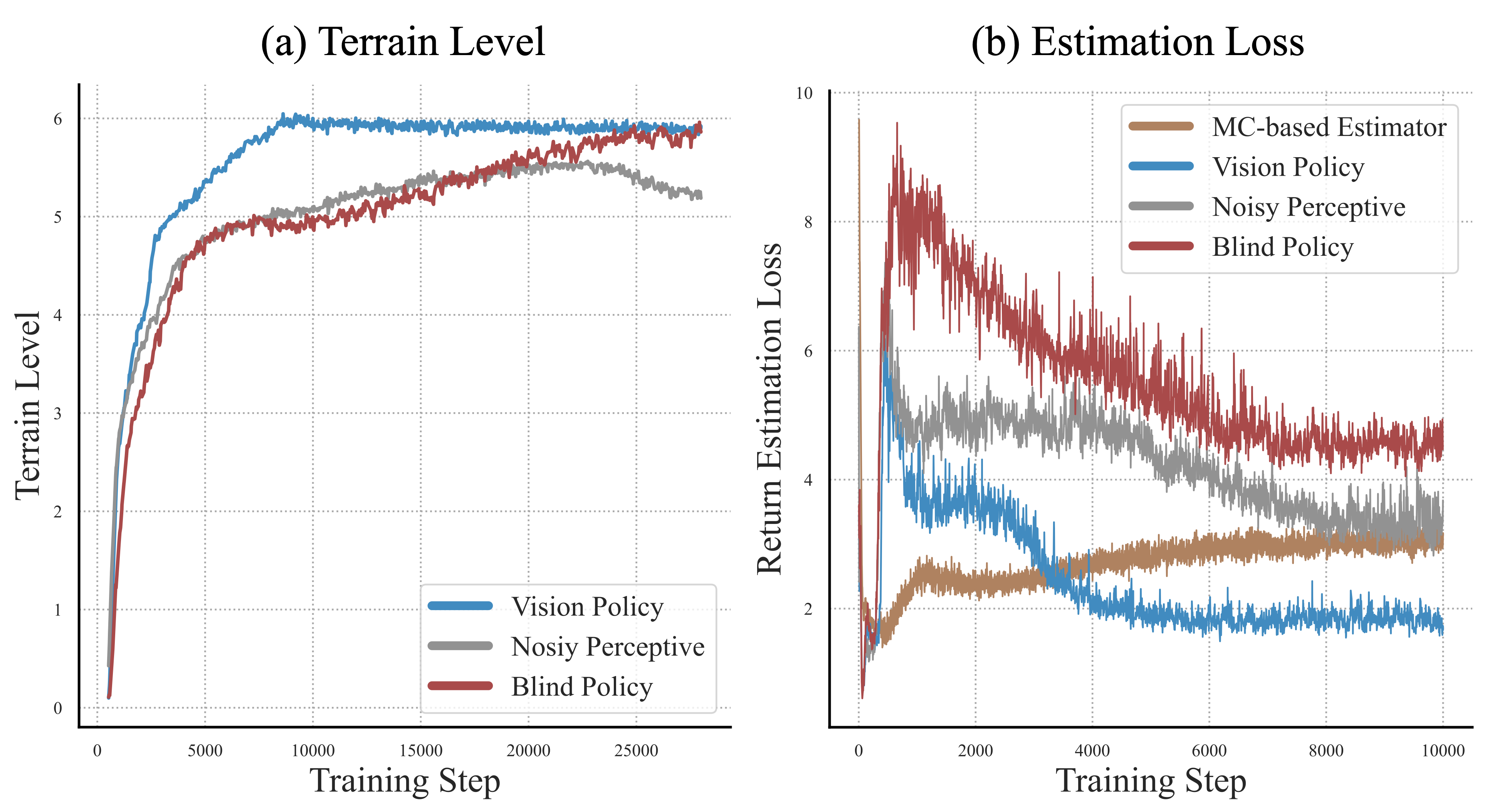}}
\caption{Training curves for terrain levels and the return estimation loss.}
\label{train}
\end{figure}

Moreover, the terrain level rises slowly for the Noisy Perceptive policy (Fig. \ref{train}-(a)), and it fails to reach the maximum level achieved by the vision and blind policies. This is because the policy struggles with the trade-off of whether to trust the external perception, which requires the addition of an extra module to address the challenge. This slow progression highlights the difficulty of handling high levels of perception deficiency, whereas VB-Com can efficiently navigate such situations by leveraging the strengths of both the vision and blind policies.

\begin{table}[!ht]
\caption{Return Estimation Evaluations}
\label{tab:RE}
\begin{center}
\renewcommand\arraystretch{1.25}
\begin{tabular}{lcccc}
\toprule[1.0pt]
Method & Goals Completed($\%$) & Collisions & Reach Steps\\
\midrule[0.8pt]

100-steps) & $78.24 \pm 1.86$ & \bm{$2.49 \pm 0.04$} & $193.7 \pm 3.2$\\  
RE(50-steps)  & \bm{$81.90 \pm 2.81$} & $2.75 \pm 0.17$ & $184.6 \pm 1.4$\\ 
Re(5-steps)   & $69.90 \pm 7.34$ & $5.23 \pm 0.59$ & $192.6 \pm 3.3$\\  
Re(1-step)    & $59.57 \pm 2.00$ & $4.78 \pm 0.16$ & \bm{$167.4 \pm 5.0$}\\  
MC-based      & $74.14 \pm 2.69$ & $4.26 \pm 0.56$ & $192.8 \pm 11.8$\\  

\bottomrule[1.0pt]
\end{tabular}
\end{center}
\end{table}

\subsection{Return Estimator Evaluations}
\textbf{The proposed return estimator achieves accurate and efficient return estimation with accessible states observations.} Since we update the return estimator using temporal difference, we compare it with the Monte Carlo-based search return estimator that estimate the furtuen expected returns with the following regression loss directly: $\mathbb{E}_t[\hat{G}_{\pi_i}^e(s_t) - \sum_{t} ^ {t+T} \gamma^t r(s_t, a_t)]$. As shown in Fig. \ref{train}-(a), the MC-based estimator struggles to converge due to the accumulation of noise. In contrast, the proposed TD-based return estimator within the vision policy convergent stably as it updates alongside the locomotion policy. The results in Tab \ref{tab:RE} further highlight the ineffectiveness of the MC-based return estimator in providing accurate estimations to guide the policy composition. Specifically, the MC-based estimator struggles to respond promptly to collisions with obstacles, this delay in response leads to larger collisions and longer reach steps, as the policy cannot effectively adjust its actions in real-time. 

\textbf{We also evaluate the impact of different switch periods (T), which define the expected return duration during return estimator updates.} While training performance remains consistent across varying periods, we observe that excessively short switch periods can negatively impact system performance. In such cases, the two policies may conflict, resulting in incomplete motion trajectories when traversing the challenging terrains and failures.

\textbf{We observe that training effectiveness is highly dependent on data variance.} For instance, the estimator within vision policy converges the fastest due to its access to more accurate and comprehensive state observations, leading to fewer low-return instances. In contrast, the estimator within Noisy Perceptive and blind policies encounter more collisions and lower returns, causing their loss to degrade more slowly.

\textbf{We demonstrate that the estimated return threhold $G_{th}$ is crucial to the performance of VB-Com.} Tab \ref{tab:TH} evaluates the system's performance under different values of $\alpha$, as well as without $G_{th}$. The results demonstrate that $G_{th}$ is critical for mitigating miscorrection during motion abnormalities, and that a value of $\alpha < 1.0$ ensures a sensitive response to the states that could lead to motion failures.

\begin{table}[!ht]
\caption{Estimated Return Threhold Evaluations}
\label{tab:TH}
\begin{center}
\renewcommand\arraystretch{1.25}
\begin{tabular}{lcccc}
\toprule[1.0pt]
Method & Goals Completed($\%$) & Collisions & Reach Steps\\
\midrule[0.8pt]
 
$\alpha = 2.0$   & $77.10 \pm 4.71$ & $2.63 \pm 0.68$ & $185.11 \pm 7.17$\\ 
$\alpha = 0.5$   & \bm{$85.76 \pm 2.88$} & $2.29 \pm 0.17$ & $186.96 \pm 3.83$\\  
$\alpha = 0.1$   & $84.43 \pm 1.23$ & \bm{$2.10 \pm 0.25$} & $\bm{184.35 \pm 6.27}$\\  
w/o $G_{th}$     & $48.48 \pm 1.28$ & $6.24 \pm 0.41$ & $261.96 \pm 35.63$\\  

\bottomrule[1.0pt]
\end{tabular}
\end{center}
\end{table}

\subsection{Real-World Experiments}

We deploy the proposed system on both the Unitree G1 and Unitree H1 robots and evaluate the performance of the proposed VB-Com method. 
\begin{figure*}[h]
\centering{\includegraphics[width= \textwidth]{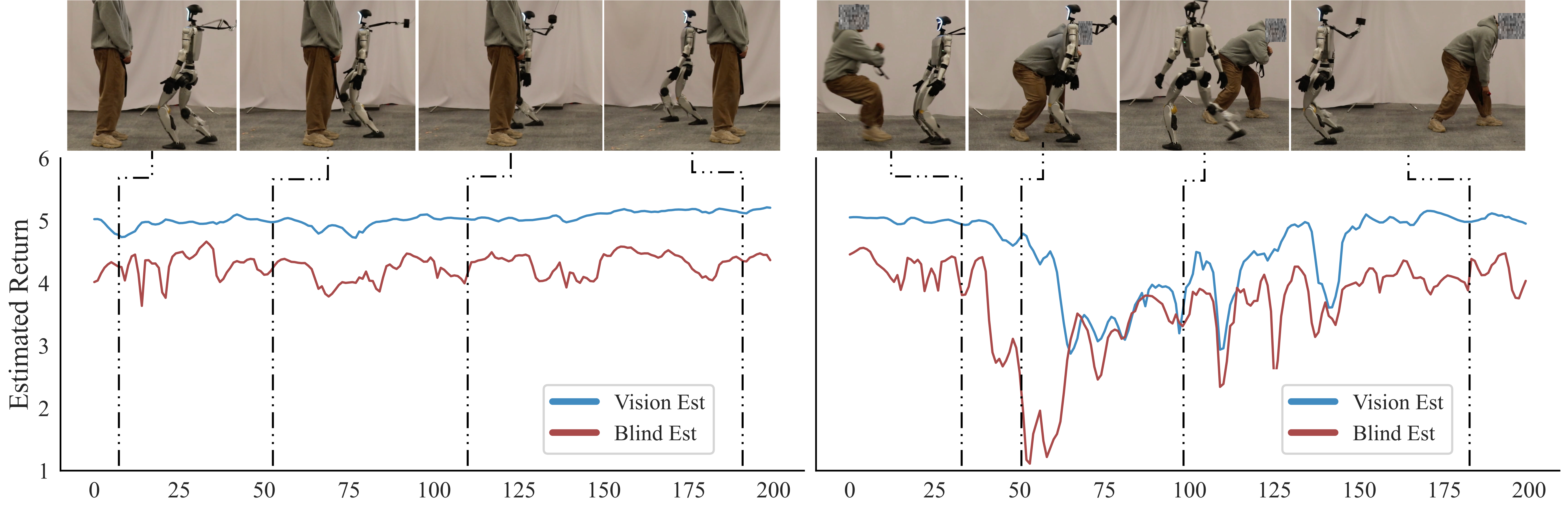}}
\caption{Illustrations of the variation in estimated return under static/dynamic obstacles in hardware experiments.}
\label{hardwarecurve}
\end{figure*}

\subsubsection{Hardware Return Estimations}

We illustrate how VB-Com operates on real robots by plotting $4$ seconds of the estimated return while the robot avoids static (left) and dynamic (right) obstacles (Fig \ref{hardwarecurve}). The results demonstrate that, for static obstacles (a standing person), the elevation map can accurately perceive the obstacle, allowing the robot to plan motions in advance and avoid collisions. Corresponding to this behavior, we observe that the estimated return on the G1 stays a high value, with $\hat{G}^e_{\pi_b}$ slightly lower than $\hat{G}^e_{\pi_v}$, consistent with the scenario where the vision policy continues to operate within VB-Com.

On the other hand, when a person moves towards the robot at high speed, the perception module fails to detect the obstacle, causing a collision, both $\hat{G}^e_{\pi_b}$ and $\hat{G}^e_{\pi_v}$ decline sharply upon collision. However, VB-Com quickly switches to $\pi_b$ to avoid the person, demonstrating the  \textbf{rapid response to collision provided by the proposed return estimation and the successful obstacle avoidance capability of VB-Com under perceptual deficiency}.

\begin{figure}[h]
\centering{\includegraphics[width=0.5\textwidth]{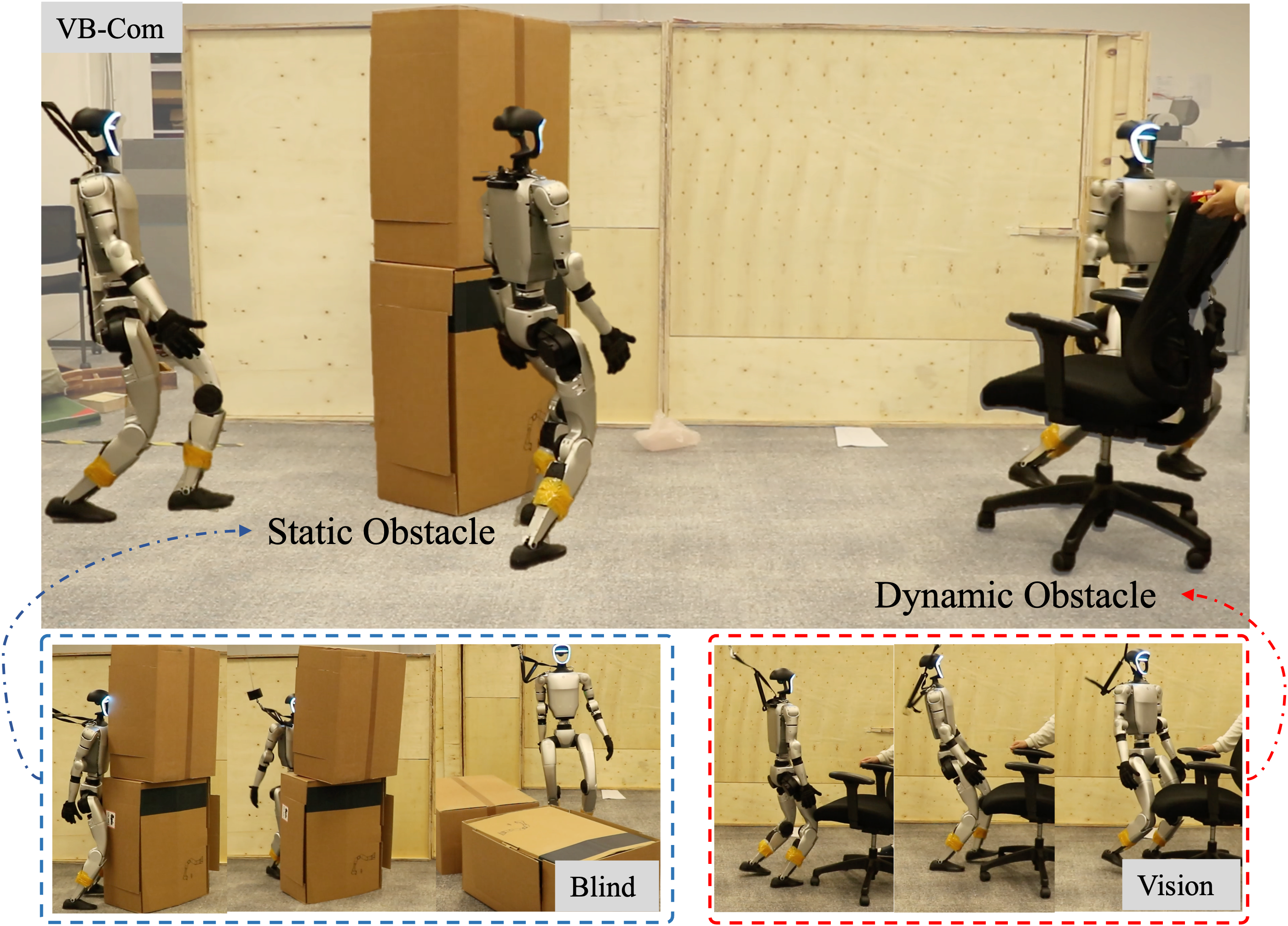}}
\caption{ Real-world comparisons of VB-Com, vision, and blind policies in obstacle avoidance on the G1.}
\label{avoid}
\end{figure}

\subsubsection{Avoid Obstacles}
In this section, we make comparisons between VB-Com along with the vision policy and blind policy on G1 (Fig \ref{avoid}), to demonstrate the superior performance of VB-Com in hardware compared with signle policies. In the evaluation scenario, G1 encounters two consecutive obstacles along its path. The second dynamic obstacle obstructs the robot's direction before the elevation map can perceive it. VB-Com enables the robot to avoid the static obstacle without collision and subsequently avoid the dynamic obstacle after it collides with the suddenly appearing obstacle.

In contrast, for the baseline policies, the blind policy makes unnecessary contact with the static obstacles before avoiding them, which damages the environment. As for the vision policy, the robot collides with the obstacle and is unable to avoid it until the newly added obstacle is detected and integrated into the map.

\begin{figure}[h]
\centering{\includegraphics[width=0.5\textwidth]{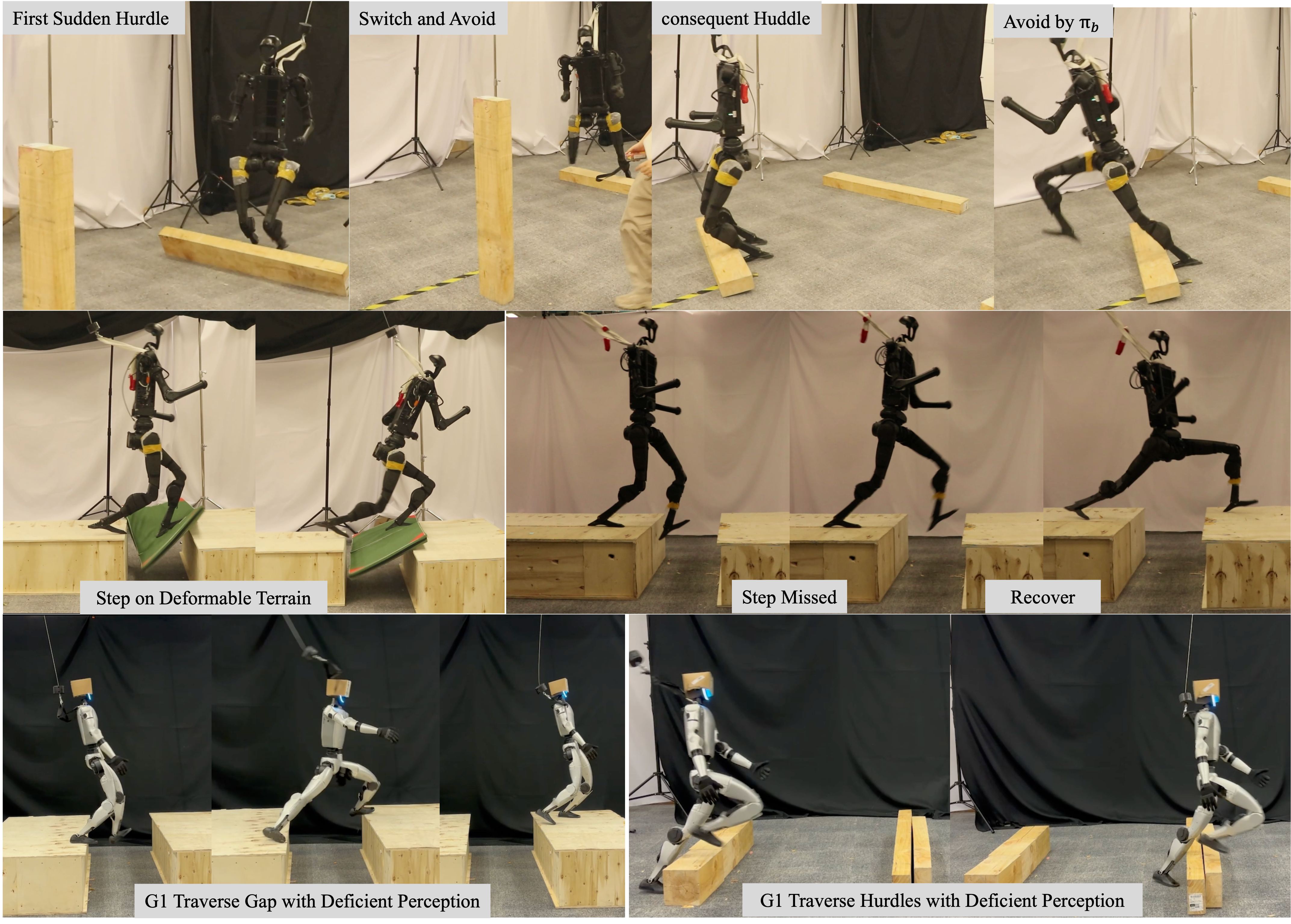}}
\caption{Hardware demonstrations on the robots traversing gaps and hurldes given deficient perception with VB-Com.}
\label{hurdlegap}
\end{figure}

\subsubsection{Performance Against Deficient Perception}
In this section, we demonstrate the ability of VB-Com to traverse challenging terrains given deficient perception (Fig. \ref{hurdlegap}). We provide zero inputs for the heightmaps to evaluate the performance of VB-Com under perceptual deficiency. We introduce two consecutive hurdles, and the robot successfully recovers after colliding with them by switching to $\pi_b$. Additionally, we demonstrate that VB-Com enables recovery from a missed step on an unobserved gap. In this case, VB-Com saves the robot by performing a larger forward step to traverse the gap without perception, as the blind policy has learned during simulation.

\section{Conclusion and Limitations} \label{sec:conclusion}

In this work, we propose VB-Com, a novel humanoid locomotion control framework that successfully combines the strengths of both vision and blind policies to enhance the performance of humanoid robots in dynamic, unstructured environments.  VB-Com achieves dynamically switching between vision and blind policies based on return estimations, enabling the robot to navigate more efficiently in complex terrains against perception. Extensive evaluations on obstacles and dynamic terrains demonstrate that VB-Com outperforms both vision-only and blind-only policies. Additionally, the system's deployment on Unitree G1 and H1 robots confirms its practical viability. This work underscores the potential of policy composition to tackle the limitations of traditional approaches and improve the robustness and adaptability of perceptive humanoid locomotion.

A key limitation of this work is that the performance upper bound is constrained by the simulated terrains encountered during training. Specifically, for unobserved gap terrains, the proposed method may fail if no suitable steppable terrain is available when the robot attempts a larger step for recovery. To address this limitation, future work will explore the incorporation of additional sub-policies to better handle such challenging scenarios and further improve the robustness of the system.

\bibliographystyle{plainnat}
\bibliography{references}

\begin{thebibliography}{49}
\providecommand{\natexlab}[1]{#1}
\providecommand{\url}[1]{\texttt{#1}}
\expandafter\ifx\csname urlstyle\endcsname\relax
  \providecommand{\doi}[1]{doi: #1}\else
  \providecommand{\doi}{doi: \begingroup \urlstyle{rm}\Url}\fi

\bibitem[Agarwal et~al.(2023)Agarwal, Kumar, Malik, and Pathak]{agarwal2023legged}
Ananye Agarwal, Ashish Kumar, Jitendra Malik, and Deepak Pathak.
\newblock Legged locomotion in challenging terrains using egocentric vision.
\newblock In \emph{Conference on robot learning}, pages 403--415. PMLR, 2023.

\bibitem[Authors(2024)]{Genesis}
Genesis Authors.
\newblock Genesis: A universal and generative physics engine for robotics and beyond, December 2024.
\newblock URL \url{https://github.com/Genesis-Embodied-AI/Genesis}.

\bibitem[Bacon et~al.(2017)Bacon, Harb, and Precup]{bacon2017option}
Pierre-Luc Bacon, Jean Harb, and Doina Precup.
\newblock The option-critic architecture.
\newblock In \emph{Proceedings of the AAAI conference on artificial intelligence}, volume~31, 2017.

\bibitem[Chen et~al.(2024{\natexlab{a}})Chen, Frey, Zhou, Miki, Martius, and Hutter]{chen2024identifying}
Jiaqi Chen, Jonas Frey, Ruyi Zhou, Takahiro Miki, Georg Martius, and Marco Hutter.
\newblock Identifying terrain physical parameters from vision-towards physical-parameter-aware locomotion and navigation.
\newblock \emph{IEEE Robotics and Automation Letters}, 2024{\natexlab{a}}.

\bibitem[Chen et~al.(2024{\natexlab{b}})Chen, He, Wang, Liao, Ze, Li, Sastry, Wu, Sreenath, Gupta, et~al.]{chen2024learning}
Zixuan Chen, Xialin He, Yen-Jen Wang, Qiayuan Liao, Yanjie Ze, Zhongyu Li, S~Shankar Sastry, Jiajun Wu, Koushil Sreenath, Saurabh Gupta, et~al.
\newblock Learning smooth humanoid locomotion through lipschitz-constrained policies.
\newblock \emph{arXiv preprint arXiv:2410.11825}, 2024{\natexlab{b}}.

\bibitem[Cheng et~al.(2024{\natexlab{a}})Cheng, Ji, Chen, Yang, Yang, and Wang]{cheng2024expressive}
Xuxin Cheng, Yandong Ji, Junming Chen, Ruihan Yang, Ge~Yang, and Xiaolong Wang.
\newblock Expressive whole-body control for humanoid robots.
\newblock \emph{arXiv preprint arXiv:2402.16796}, 2024{\natexlab{a}}.

\bibitem[Cheng et~al.(2024{\natexlab{b}})Cheng, Shi, Agarwal, and Pathak]{cheng2024extreme}
Xuxin Cheng, Kexin Shi, Ananye Agarwal, and Deepak Pathak.
\newblock Extreme parkour with legged robots.
\newblock In \emph{2024 IEEE International Conference on Robotics and Automation (ICRA)}, pages 11443--11450. IEEE, 2024{\natexlab{b}}.

\bibitem[Choi et~al.(2023)Choi, Ji, Park, Kim, Mun, Lee, and Hwangbo]{choi2023learning}
Suyoung Choi, Gwanghyeon Ji, Jeongsoo Park, Hyeongjun Kim, Juhyeok Mun, Jeong~Hyun Lee, and Jemin Hwangbo.
\newblock Learning quadrupedal locomotion on deformable terrain.
\newblock \emph{Science Robotics}, 8\penalty0 (74):\penalty0 eade2256, 2023.

\bibitem[Cui et~al.()Cui, Li, Huang, Qin, Zhang, Zheng, Tang, Hu, Yan, Chen, et~al.]{cuiadapting}
Wenhao Cui, Shengtao Li, Huaxing Huang, Bangyu Qin, Tianchu Zhang, Liang Zheng, Ziyang Tang, Chenxu Hu, NING Yan, Jiahao Chen, et~al.
\newblock Adapting humanoid locomotion over challenging terrain via two-phase training.
\newblock In \emph{8th Annual Conference on Robot Learning}.

\bibitem[Fu et~al.(2022)Fu, Kumar, Agarwal, Qi, Malik, and Pathak]{fu2022coupling}
Zipeng Fu, Ashish Kumar, Ananye Agarwal, Haozhi Qi, Jitendra Malik, and Deepak Pathak.
\newblock Coupling vision and proprioception for navigation of legged robots.
\newblock In \emph{Proceedings of the IEEE/CVF Conference on Computer Vision and Pattern Recognition}, pages 17273--17283, 2022.

\bibitem[Fu et~al.(2024)Fu, Zhao, Wu, Wetzstein, and Finn]{fu2024humanplus}
Zipeng Fu, Qingqing Zhao, Qi~Wu, Gordon Wetzstein, and Chelsea Finn.
\newblock Humanplus: Humanoid shadowing and imitation from humans.
\newblock \emph{arXiv preprint arXiv:2406.10454}, 2024.

\bibitem[Gu et~al.(2024{\natexlab{a}})Gu, Wang, and Chen]{gu2024humanoid}
Xinyang Gu, Yen-Jen Wang, and Jianyu Chen.
\newblock Humanoid-gym: Reinforcement learning for humanoid robot with zero-shot sim2real transfer.
\newblock \emph{arXiv preprint arXiv:2404.05695}, 2024{\natexlab{a}}.

\bibitem[Gu et~al.(2024{\natexlab{b}})Gu, Wang, Zhu, Shi, Guo, Liu, and Chen]{gu2024advancing}
Xinyang Gu, Yen-Jen Wang, Xiang Zhu, Chengming Shi, Yanjiang Guo, Yichen Liu, and Jianyu Chen.
\newblock Advancing humanoid locomotion: Mastering challenging terrains with denoising world model learning.
\newblock \emph{arXiv preprint arXiv:2408.14472}, 2024{\natexlab{b}}.

\bibitem[Gupta et~al.(2023)Gupta, Lynch, Kinman, Peake, Levine, and Hausman]{gupta2023bootstrapped}
Abhishek Gupta, Corey Lynch, Brandon Kinman, Garrett Peake, Sergey Levine, and Karol Hausman.
\newblock Bootstrapped autonomous practicing via multi-task reinforcement learning.
\newblock In \emph{2023 IEEE International Conference on Robotics and Automation (ICRA)}, pages 5020--5026. IEEE, 2023.

\bibitem[He et~al.(2024{\natexlab{a}})He, Luo, He, Xiao, Zhang, Zhang, Kitani, Liu, and Shi]{he2024omnih2o}
Tairan He, Zhengyi Luo, Xialin He, Wenli Xiao, Chong Zhang, Weinan Zhang, Kris Kitani, Changliu Liu, and Guanya Shi.
\newblock Omnih2o: Universal and dexterous human-to-humanoid whole-body teleoperation and learning.
\newblock \emph{arXiv preprint arXiv:2406.08858}, 2024{\natexlab{a}}.

\bibitem[He et~al.(2024{\natexlab{b}})He, Xiao, Lin, Luo, Xu, Jiang, Kautz, Liu, Shi, Wang, et~al.]{he2024hover}
Tairan He, Wenli Xiao, Toru Lin, Zhengyi Luo, Zhenjia Xu, Zhenyu Jiang, Jan Kautz, Changliu Liu, Guanya Shi, Xiaolong Wang, et~al.
\newblock Hover: Versatile neural whole-body controller for humanoid robots.
\newblock \emph{arXiv preprint arXiv:2410.21229}, 2024{\natexlab{b}}.

\bibitem[He et~al.(2024{\natexlab{c}})He, Zhang, Xiao, He, Liu, and Shi]{he2024agile}
Tairan He, Chong Zhang, Wenli Xiao, Guanqi He, Changliu Liu, and Guanya Shi.
\newblock Agile but safe: Learning collision-free high-speed legged locomotion.
\newblock \emph{arXiv preprint arXiv:2401.17583}, 2024{\natexlab{c}}.

\bibitem[Hoeller et~al.(2024)Hoeller, Rudin, Sako, and Hutter]{hoeller2024anymal}
David Hoeller, Nikita Rudin, Dhionis Sako, and Marco Hutter.
\newblock Anymal parkour: Learning agile navigation for quadrupedal robots.
\newblock \emph{Science Robotics}, 9\penalty0 (88):\penalty0 eadi7566, 2024.

\bibitem[Hwangbo et~al.(2019)Hwangbo, Lee, Dosovitskiy, Bellicoso, Tsounis, Koltun, and Hutter]{hwangbo2019learning}
Jemin Hwangbo, Joonho Lee, Alexey Dosovitskiy, Dario Bellicoso, Vassilios Tsounis, Vladlen Koltun, and Marco Hutter.
\newblock Learning agile and dynamic motor skills for legged robots.
\newblock \emph{Science Robotics}, 4\penalty0 (26):\penalty0 eaau5872, 2019.

\bibitem[Ji et~al.(2024)Ji, Peng, Liu, Li, Yang, Cheng, and Wang]{ji2024exbody2}
Mazeyu Ji, Xuanbin Peng, Fangchen Liu, Jialong Li, Ge~Yang, Xuxin Cheng, and Xiaolong Wang.
\newblock Exbody2: Advanced expressive humanoid whole-body control.
\newblock \emph{arXiv preprint arXiv:2412.13196}, 2024.

\bibitem[Kumar et~al.(2021)Kumar, Fu, Pathak, and Malik]{kumar2021rma}
Ashish Kumar, Zipeng Fu, Deepak Pathak, and Jitendra Malik.
\newblock Rma: Rapid motor adaptation for legged robots.
\newblock \emph{arXiv preprint arXiv:2107.04034}, 2021.

\bibitem[Lee et~al.(2020)Lee, Hwangbo, Wellhausen, Koltun, and Hutter]{lee2020learning}
Joonho Lee, Jemin Hwangbo, Lorenz Wellhausen, Vladlen Koltun, and Marco Hutter.
\newblock Learning quadrupedal locomotion over challenging terrain.
\newblock \emph{Science robotics}, 5\penalty0 (47):\penalty0 eabc5986, 2020.

\bibitem[Li et~al.(2023)Li, Peng, Abbeel, Levine, Berseth, and Sreenath]{li2023robust}
Zhongyu Li, Xue~Bin Peng, Pieter Abbeel, Sergey Levine, Glen Berseth, and Koushil Sreenath.
\newblock Robust and versatile bipedal jumping control through reinforcement learning.
\newblock \emph{arXiv preprint arXiv:2302.09450}, 2023.

\bibitem[Long et~al.(2024{\natexlab{a}})Long, Ren, Shi, Wang, Huang, Luo, and Pang]{long2024learning}
Junfeng Long, Junli Ren, Moji Shi, Zirui Wang, Tao Huang, Ping Luo, and Jiangmiao Pang.
\newblock Learning humanoid locomotion with perceptive internal model.
\newblock \emph{arXiv preprint arXiv:2411.14386}, 2024{\natexlab{a}}.

\bibitem[Long et~al.(2024{\natexlab{b}})Long, Wang, Li, Cao, Gao, and Pang]{long2024hybrid}
Junfeng Long, Zirui Wang, Quanyi Li, Liu Cao, Jiawei Gao, and Jiangmiao Pang.
\newblock Hybrid internal model: Learning agile legged locomotion with simulated robot response.
\newblock In \emph{The Twelfth International Conference on Learning Representations}, 2024{\natexlab{b}}.

\bibitem[Lu et~al.(2024)Lu, Cheng, Li, Yang, Ji, Yuan, Yang, Yi, and Wang]{lu2024mobile}
Chenhao Lu, Xuxin Cheng, Jialong Li, Shiqi Yang, Mazeyu Ji, Chengjing Yuan, Ge~Yang, Sha Yi, and Xiaolong Wang.
\newblock Mobile-television: Predictive motion priors for humanoid whole-body control.
\newblock \emph{arXiv preprint arXiv:2412.07773}, 2024.

\bibitem[Luo et~al.(2024)Luo, Li, Yu, Wang, Wu, and Zhu]{luo2024pie}
Shixin Luo, Songbo Li, Ruiqi Yu, Zhicheng Wang, Jun Wu, and Qiuguo Zhu.
\newblock Pie: Parkour with implicit-explicit learning framework for legged robots.
\newblock \emph{IEEE Robotics and Automation Letters}, 2024.

\bibitem[Makoviychuk et~al.(2021)Makoviychuk, Wawrzyniak, Guo, Lu, Storey, Macklin, Hoeller, Rudin, Allshire, Handa, et~al.]{makoviychuk2021isaac}
Viktor Makoviychuk, Lukasz Wawrzyniak, Yunrong Guo, Michelle Lu, Kier Storey, Miles Macklin, David Hoeller, Nikita Rudin, Arthur Allshire, Ankur Handa, et~al.
\newblock Isaac gym: High performance gpu-based physics simulation for robot learning.
\newblock \emph{arXiv preprint arXiv:2108.10470}, 2021.

\bibitem[Margolis and Agrawal(2023)]{margolis2023walk}
Gabriel~B Margolis and Pulkit Agrawal.
\newblock Walk these ways: Tuning robot control for generalization with multiplicity of behavior.
\newblock In \emph{Conference on Robot Learning}, pages 22--31. PMLR, 2023.

\bibitem[Margolis et~al.(2024)Margolis, Yang, Paigwar, Chen, and Agrawal]{margolis2024rapid}
Gabriel~B Margolis, Ge~Yang, Kartik Paigwar, Tao Chen, and Pulkit Agrawal.
\newblock Rapid locomotion via reinforcement learning.
\newblock \emph{The International Journal of Robotics Research}, 43\penalty0 (4):\penalty0 572--587, 2024.

\bibitem[Miki et~al.(2022{\natexlab{a}})Miki, Lee, Hwangbo, Wellhausen, Koltun, and Hutter]{miki2022learning}
Takahiro Miki, Joonho Lee, Jemin Hwangbo, Lorenz Wellhausen, Vladlen Koltun, and Marco Hutter.
\newblock Learning robust perceptive locomotion for quadrupedal robots in the wild.
\newblock \emph{Science robotics}, 7\penalty0 (62):\penalty0 eabk2822, 2022{\natexlab{a}}.

\bibitem[Miki et~al.(2022{\natexlab{b}})Miki, Wellhausen, Grandia, Jenelten, Homberger, and Hutter]{miki2022elevation}
Takahiro Miki, Lorenz Wellhausen, Ruben Grandia, Fabian Jenelten, Timon Homberger, and Marco Hutter.
\newblock Elevation mapping for locomotion and navigation using gpu.
\newblock In \emph{2022 IEEE/RSJ International Conference on Intelligent Robots and Systems (IROS)}, pages 2273--2280. IEEE, 2022{\natexlab{b}}.

\bibitem[Nahrendra et~al.(2023)Nahrendra, Yu, and Myung]{nahrendra2023dreamwaq}
I~Made~Aswin Nahrendra, Byeongho Yu, and Hyun Myung.
\newblock Dreamwaq: Learning robust quadrupedal locomotion with implicit terrain imagination via deep reinforcement learning.
\newblock In \emph{2023 IEEE International Conference on Robotics and Automation (ICRA)}, pages 5078--5084. IEEE, 2023.

\bibitem[Nasiriany et~al.(2022)Nasiriany, Liu, and Zhu]{nasiriany2022augmenting}
Soroush Nasiriany, Huihan Liu, and Yuke Zhu.
\newblock Augmenting reinforcement learning with behavior primitives for diverse manipulation tasks.
\newblock In \emph{2022 International Conference on Robotics and Automation (ICRA)}, pages 7477--7484. IEEE, 2022.

\bibitem[Peng et~al.(2019)Peng, Chang, Zhang, Abbeel, and Levine]{peng2019mcp}
Xue~Bin Peng, Michael Chang, Grace Zhang, Pieter Abbeel, and Sergey Levine.
\newblock Mcp: Learning composable hierarchical control with multiplicative compositional policies.
\newblock \emph{Advances in neural information processing systems}, 32, 2019.

\bibitem[Radosavovic et~al.(2024)Radosavovic, Kamat, Darrell, and Malik]{radosavovic2024learning}
Ilija Radosavovic, Sarthak Kamat, Trevor Darrell, and Jitendra Malik.
\newblock Learning humanoid locomotion over challenging terrain.
\newblock \emph{arXiv preprint arXiv:2410.03654}, 2024.

\bibitem[Ren et~al.(2024)Ren, Liu, Dai, Long, and Wang]{ren2024top}
Junli Ren, Yikai Liu, Yingru Dai, Junfeng Long, and Guijin Wang.
\newblock Top-nav: Legged navigation integrating terrain, obstacle and proprioception estimation.
\newblock \emph{arXiv preprint arXiv:2404.15256}, 2024.

\bibitem[Schulman et~al.(2015)Schulman, Moritz, Levine, Jordan, and Abbeel]{schulman2015high}
John Schulman, Philipp Moritz, Sergey Levine, Michael Jordan, and Pieter Abbeel.
\newblock High-dimensional continuous control using generalized advantage estimation.
\newblock \emph{arXiv preprint arXiv:1506.02438}, 2015.

\bibitem[Shah et~al.(2021)Shah, Xu, Lu, Xiao, Toshev, Levine, and Ichter]{shah2021value}
Dhruv Shah, Peng Xu, Yao Lu, Ted Xiao, Alexander Toshev, Sergey Levine, and Brian Ichter.
\newblock Value function spaces: Skill-centric state abstractions for long-horizon reasoning.
\newblock \emph{arXiv preprint arXiv:2111.03189}, 2021.

\bibitem[Sun et~al.(2024)Sun, Zhou, Geng, Zhang, and Li]{sun2024leg}
Jingyu Sun, Lelai Zhou, Binghou Geng, Yi~Zhang, and Yibin Li.
\newblock Leg state estimation for quadruped robot by using probabilistic model with proprioceptive feedback.
\newblock \emph{IEEE/ASME transactions on mechatronics}, 2024.

\bibitem[Yang et~al.(2021)Yang, Zhang, Hansen, Xu, and Wang]{yang2021learning}
Ruihan Yang, Minghao Zhang, Nicklas Hansen, Huazhe Xu, and Xiaolong Wang.
\newblock Learning vision-guided quadrupedal locomotion end-to-end with cross-modal transformers.
\newblock \emph{arXiv preprint arXiv:2107.03996}, 2021.

\bibitem[Yang et~al.(2023)Yang, Yang, and Wang]{yang2023neural}
Ruihan Yang, Ge~Yang, and Xiaolong Wang.
\newblock Neural volumetric memory for visual locomotion control.
\newblock In \emph{Proceedings of the IEEE/CVF Conference on Computer Vision and Pattern Recognition}, pages 1430--1440, 2023.

\bibitem[Yu et~al.(2024)Yu, Wang, Wang, Wang, Wu, and Zhu]{yu2024walking}
Ruiqi Yu, Qianshi Wang, Yizhen Wang, Zhicheng Wang, Jun Wu, and Qiuguo Zhu.
\newblock Walking with terrain reconstruction: Learning to traverse risky sparse footholds.
\newblock \emph{arXiv preprint arXiv:2409.15692}, 2024.

\bibitem[Zakka et~al.(2025)Zakka, Tabanpour, Liao, Haiderbhai, Holt, Luo, Allshire, Frey, Sreenath, Kahrs, et~al.]{zakka2025mujoco}
Kevin Zakka, Baruch Tabanpour, Qiayuan Liao, Mustafa Haiderbhai, Samuel Holt, Jing~Yuan Luo, Arthur Allshire, Erik Frey, Koushil Sreenath, Lueder~A Kahrs, et~al.
\newblock Mujoco playground.
\newblock \emph{arXiv preprint arXiv:2502.08844}, 2025.

\bibitem[Zhang et~al.(2024)Zhang, Jin, Frey, Rudin, Mattamala, Cadena, and Hutter]{zhang2024resilient}
Chong Zhang, Jin Jin, Jonas Frey, Nikita Rudin, Mat{\'\i}as Mattamala, Cesar Cadena, and Marco Hutter.
\newblock Resilient legged local navigation: Learning to traverse with compromised perception end-to-end.
\newblock In \emph{2024 IEEE International Conference on Robotics and Automation (ICRA)}, pages 34--41. IEEE, 2024.

\bibitem[Zhang et~al.(2023)Zhang, Xu, and Yu]{zhang2023policy}
Haichao Zhang, We~Xu, and Haonan Yu.
\newblock Policy expansion for bridging offline-to-online reinforcement learning.
\newblock \emph{arXiv preprint arXiv:2302.00935}, 2023.

\bibitem[Zhu et~al.(2025)Zhu, Mou, Li, Ye, Huang, and Zhao]{zhu2025vr}
Shaoting Zhu, Linzhan Mou, Derun Li, Baijun Ye, Runhan Huang, and Hang Zhao.
\newblock Vr-robo: A real-to-sim-to-real framework for visual robot navigation and locomotion.
\newblock \emph{arXiv preprint arXiv:2502.01536}, 2025.

\bibitem[Zhuang et~al.(2023)Zhuang, Fu, Wang, Atkeson, Schwertfeger, Finn, and Zhao]{zhuang2023robot}
Ziwen Zhuang, Zipeng Fu, Jianren Wang, Christopher Atkeson, Soeren Schwertfeger, Chelsea Finn, and Hang Zhao.
\newblock Robot parkour learning.
\newblock \emph{arXiv preprint arXiv:2309.05665}, 2023.

\bibitem[Zhuang et~al.(2024)Zhuang, Yao, and Zhao]{zhuang2024humanoid}
Ziwen Zhuang, Shenzhe Yao, and Hang Zhao.
\newblock Humanoid parkour learning.
\newblock \emph{arXiv preprint arXiv:2406.10759}, 2024.

\end{thebibliography}

\clearpage
\newpage
\onecolumn
\begin{appendices}
\section{Reward Functions}
The reward functions used during training for both the H1 and G1 are shown in Table \ref{reward}. We categorize the rewards into three types: task rewards, which guide the robot in tracking goal points and avoiding collisions (important for obstacle avoidance); regularization rewards, which impose constraints on smooth motion and hardware protection; and motion style rewards, which enforce constraints for a human-like whole-body motion style. The target base height $h^{\text {target}}$ for G1 is set to $0.728m$, The minimum feet/knee lateral distance for G1 and H1 is set to $d_\text{min}^{g1} = 0.18m$ and $d_\text{min}^{h1} = 0.25m$.
\begin{table*}[!h]
    \centering
    \caption{Rewards}
    \begin{tabular}{llll}
    \toprule[1.5pt] Reward & Equation & Weight: H1 & Weight: G1 \\ \midrule[1.5pt] 
    \multicolumn{4}{c}{\textbf{Task Rewards}} \\ [0.4ex]
    Tracking Goal Velocity & $\min(v_{c},\|\mathbf{v}_{x y}\|)/v_{c}$ & 5.0 & 2.0 \\[0.4ex]
    Tracking Yaw & $\exp \left\{(\textbf{p} - \textbf{x}) / ||\textbf{p} - \textbf{x}|| \right\}$ & 5.0 & 2.0\\[0.4ex]
    Collision & $\sum_{i\in C_l^e}{\bm{1}}\left\{\|\mathbf{f}_i \|>0.1\right \}$ & -15.0 & -15.0 \\[0.4ex]
    \toprule[1.0pt] \multicolumn{4}{c}{\textbf{Regularization Rewards}} \\ [0.4ex]
    Linear velocity $(z)$ & $v_z^2$ & -1.0 & -1.0\\ [0.4ex]
    Angular velocity $(x y)$ & $\|\boldsymbol{\omega}_{x y}\|_2^2$ & -0.05 & -0.05\\[0.4ex]
    Orientation & $\|\mathbf{g}_{x}\|_2^2 + \|\mathbf{g}_{y}\|_2^2$ & -2.0 & -2.0\\[0.4ex]
    Joint accelerations & $\|\ddot{\boldsymbol{\theta}}\|_2^2$ & $-2.5 \times 10^{-7}$ & $-2.5 \times 10^{-7}$\\[0.4ex]
    Joint velocity & $\|\dot{\boldsymbol{\theta}}\|_{2}^{2}$ & $-5.0 \times 10^{-4}$ & $-5.0 \times 10^{-4}$\\ [0.4ex]
    Torques & $\|\frac{\boldsymbol{\tau}}{k_p}\|_{2}^{2}$ & $-1.0 \times 10^{-5}$ & $-1.0 \times 10^{-5}$\\[0.4ex]
    Action rate & $\|\mathbf{a}_t-\mathbf{a}_{t-1}\|_2^2$ & -0.3 & -0.3 \\[0.4ex]
    Joint pos limits & $\text{RELU}(\boldsymbol{\theta} - \boldsymbol{\theta}^{max}) + \text{RELU}(\boldsymbol{\theta}^{min} - \boldsymbol{\theta})$ & -2.0 & -2.0\\[0.4ex]
    Joint vel limits & $\text{RELU}(\boldsymbol{\hat{\theta}} - \boldsymbol{\hat{\theta}}^{max})$ & -1.0 & -1.0\\[0.4ex]
    Torque limits & $\text{RELU}(\boldsymbol{\hat{\tau}} - \boldsymbol{\hat{\tau}}^{max})$ & -1.0 & -1.0 \\[0.4ex]
    \toprule[1.0pt] \multicolumn{4}{c}{\textbf{Motion Style Rewards}} \\ [0.4ex]
    Base Height & $\left(h - h^{\text {target}}\right)^2$ & -0.0 & -10.0 \\ [0.2ex]
    Feet Air Time & $\sum_{i\in \text{feet}}^2 \left( t_{\text{air}, i} - 0.5 \right) \cdot \mathbf{1}\left\{\text{first ground contact} \right\}$ & 4.0 & 1.0 \\[0.4ex]
    Feet Stumble & $\sum_{i\in \text{feet}}\mathbf{1}\left\{\left|\boldsymbol{f}_i^{x y}\right|>3\left|\boldsymbol{f}_i^z\right|\right\}$ & -1.0 & -1.0 \\[0.4ex]
    Arm joint deviation & $\sum_{i \in \text{arm}}|\boldsymbol{\theta}_{i} - \boldsymbol{\theta}^{\text{default}}_{i}|^{2}$ & -0.5 & -0.5\\[0.4ex]
    Hip joint deviation & $\sum_{i \in \text{hip}}|\boldsymbol{\theta}{i} - \boldsymbol{\theta}^{\text{default}}_{i}|^{2}$ & -5.0 & -0.5\\[0.4ex]
    Waist joint deviation & $\sum_{i \in \text{waist}}|\boldsymbol{\theta}{i} - \boldsymbol{\theta}^{\text{default}}_{i}|^{2}$ & -5.0 & -0.0\\[0.4ex]
    Feet distance & $\left(\|\mathbf{p}_\text{left foot} - \mathbf{p}_\text{left foot}\| - d_\text{min} \right)$ & 1.0 & 0.0\\[0.4ex]
    Feet lateral distance & $\left(\|\mathbf{p}^y_\text{left foot} - \mathbf{p}^y_\text{right foot}\| - d_\text{min} \right)$ & 10.0 & 0.5\\[0.4ex]
    Knee lateral distance & $\left(\|\mathbf{p}^y_\text{left knee} - \mathbf{p}^y_\text{right knee}\| - d_\text{min} \right)$ & 5.0 & 0.0\\[0.4ex]
    Feet ground parallel & $\sum_{i\in\text{feet}}\text{Var}(\boldsymbol{p}_i^z)$ & -10.0 & -0.02 \\[0.4ex]
    \bottomrule[1.5pt]
    \end{tabular}
    \label{reward}
\end{table*}
\section{Implementation Details}
\begin{table}[!htbp]
    \centering
    \caption{Implementation Details}
    \begin{tabular}{ll}
    \toprule[1.5pt]
    \textbf{Networks} & \textbf{Hiddern Layers}\\
    Actor & [512, 256, 128] \\
    Critic & [512, 256, 128] \\
    Return Estimator & [512, 256, 128] \\
    Vel Estimator & [512, 256, 32] \\
    Terrain Encoder & [128, 128, 64] \\
    \midrule
    \textbf{Hyperparameters} & \textbf{Values}\\
    Heightmap Range Forward (m) & $[-0.35, 0.85]$ \\
    Heightmap Range Lateral (m) & $[-0.35, 0.35]$ \\
    Velocity Command Range (m/s) & $[0.0, 1.0]$ \\
    Yaw Command Range (rad/s) & $[-0.5, 0.5]$ \\
    \midrule
    \textbf{Curriculum} & \textbf{Ranges} (TL: Terrain Level)\\
    Gap Width Curriculum Range (m) & $[0.1+0.5*TL, 0.2 + 0.6*TL]$ \\
    Hurdles Heights Curriculum Range (m) & $[0.1+0.1*TL, 0.2 + 0.2*TL]$ \\
    Obstacles Length Curriculum Range (m) & $[0.1+0.1*TL, 0.2 + 0.2*TL]$ \\
    \bottomrule[1.5pt]
    \end{tabular}
    \label{tab:ImplementationDetails}
\end{table}




\end{appendices}

\end{document}